%% file: main.tex
\documentclass[journal]{IEEEtran} 
\usepackage{amsmath,amsfonts}
\usepackage{algorithmic}
\usepackage{algorithm}
\usepackage{array}
\usepackage{textcomp}
\usepackage{stfloats}
\usepackage{url}
\usepackage{verbatim}
\usepackage{graphicx}
\usepackage{cite}
\usepackage{tabularx}
\usepackage{url}
\usepackage{float}
\usepackage{subcaption}

\usepackage{multirow}
\usepackage{booktabs}
\usepackage{siunitx}
\usepackage{array}
\usepackage{hyperref}

\usepackage{todonotes}
\usepackage[inline]{enumitem}

\usepackage{cleveref}
\DeclareMathOperator*{\argmin}{arg\,min}
\newcommand{\mat}[1]{\boldsymbol{#1}}
\renewcommand{\vec}[1]{\boldsymbol{#1}}

\usepackage{fancyhdr}
\fancypagestyle{mahmood}{%
  \fancyhf{} 
  
  \fancyhead[C]{\footnotesize \textcopyright 2024 IEEE.  Personal use of this material is permitted.  Permission from IEEE must be obtained for all other uses, in any current or future media, including reprinting/republishing this material for advertising or promotional purposes, creating new collective works, for resale or redistribution to servers or lists, or reuse of any copyrighted component of this work in other works.}
}%
\makeatletter
\let\ps@IEEEtitlepagestyle\ps@mahmood
\makeatother
\begin{document}

\title{Fully Onboard SLAM for Distributed Mapping \\with a Swarm of Nano-Drones}

\author{Carl Friess, Vlad Niculescu, Tommaso Polonelli, \IEEEmembership{Member, IEEE}, Michele Magno, \IEEEmembership{Senior Member, IEEE} and Luca Benini, \IEEEmembership{Fellow, IEEE}
\thanks{All authors are with the Department of Information Technology and Electrical Engineering (D-ITET), ETH Z\"urich, Switzerland}
\thanks{L. Benini is also with the Department of Electrical, Electronic and Information Engineering (DEI), University of Bologna, Italy}
\thanks{V. Niculescu is the corresponding author: {\tt\small vladn@ethz.ch}}
\thanks{Digital Object Identifier 10.1109/JIOT.2024.3367451}
}




\maketitle

\input{sources/0-abstract.tex}

\begin{IEEEkeywords}
UAV, Nano-drone, Mapping, SLAM, Swarm intelligence
\end{IEEEkeywords}

\section*{Supplementary Material}
\begin{center}
Demo video at: \\ \href{https://youtu.be/c9hajp_43aw}{\url{youtu.be/c9hajp_43aw}} \\
Code available at: \\ \href{https://github.com/ETH-PBL/Nano_Swarm_Mapping}{\url{github.com/ETH-PBL/Nano_Swarm_Mapping}}
\end{center}
\input{sources/1-introduction.tex}
\input{sources/2-related.tex}
\input{sources/3-background.tex}
\input{sources/4-algorithms.tex}
\input{sources/5-swarm.tex}
\input{sources/6-results.tex}

\section{Conclusion}
This paper presented a complete system to enable mapping on a swarm of IoRT nano-UAVs based on a novel ISCC system and four low-power multi-zone ToF sensors.
The main contributions include a novel lightweight sensor module that provides a $360^{\circ}$ depth map with an accuracy comparable with SoA LiDARs. 
Moreover, ICP and SLAM are optimized to run on resource-constrained MCUs with a total available memory of only \qty{50}{\kilo\byte}, supporting and combining sparse measurements from up to 20 swarm agents. 
Using a different wireless physical link, e.g., the UWB protocol commonly used for indoor localization,  the swarm can potentially be extended to hundreds of nano-UAVs~\cite{koul2022data}. We also demonstrated, with three field experiments, that a collaborative group of robots can accurately map an environment and decrease the mission latency linearly w.r.t. the number of employed swarm agents.
The swarm redundancy, and its improved robustness, are assured by a system-level strategy specifically designed for this work. 
Indeed, all the collected poses are broadcast to the whole swarm; thus, the \emph{main drone} can be selected at will or easily replaced by another nano-UAV without losing any information -- the swarm data is distributed across all its agents. 
Despite its extremely lightweight setup (\qty{38.4}{\gram}), the system proposed in this paper features a mapping accuracy comparable with the SoA contributions designed for MAVs and standard-size UAVs.
We showed how our swarm-based solution reduces the pose estimation and mapping errors by about half while flying with a speed of $\SI{0.8}{\meter / \second}$.
The system proposed in this paper paves the way for more autonomous micro-robots characterized by ultra-constrained hardware, \textit{de-facto} enabling a system technology not present in the nano-UAV field so far. Indeed, with a complete environmental map, advanced navigation solutions can be inferred to enhance flight autonomy with advanced path planning and mission strategies. 

\section*{Acknowledgments}
This work has been supported in part by “TinyTrainer” project that receives funding from the Swiss National Science Foundation under grant number 207913.

\bibliographystyle{IEEEtran}
\bibliography{references}

\end{document}

%% file: sources/0-abstract.tex
\begin{abstract}
The use of Unmanned Aerial Vehicles (UAVs) is rapidly increasing in applications ranging from surveillance and first-aid missions to industrial automation involving cooperation with other machines or humans. 
To maximize area coverage and reduce mission latency, swarms of collaborating drones have become a significant research direction. 
However, this approach requires open challenges in positioning, mapping, and communications to be addressed. This work describes a distributed mapping system based on a swarm of nano-UAVs, characterized by a limited payload of 35~g and tightly constrained onboard sensing and computing capabilities. Each nano-UAV is equipped with four 64-pixel depth sensors that measure the relative distance to obstacles in four directions. The proposed system merges the information from the swarm and generates a coherent grid map without relying on any external infrastructure. 
The data fusion is performed using the iterative closest point algorithm and a graph-based simultaneous localization and mapping algorithm, running entirely onboard the UAV's low-power ARM \mbox{Cortex-M} microcontroller with just 192~kB of memory. Field results gathered in three different mazes with a swarm of up to 4 nano-UAVs prove a mapping accuracy of 12~cm and demonstrate that the mapping time is inversely proportional to the number of agents. The proposed framework scales linearly in terms of communication bandwidth and onboard computational complexity, supporting communication between up to 20 nano-UAVs and mapping of areas up to 180~m\textsuperscript{2} with the chosen configuration requiring only 50~kB of memory.
\end{abstract}

%% file: sources/1-introduction.tex
\section{Introduction}
\label{sec:intro}
\begin{figure} [t]
\begin{centering}
\includegraphics[width=\columnwidth]{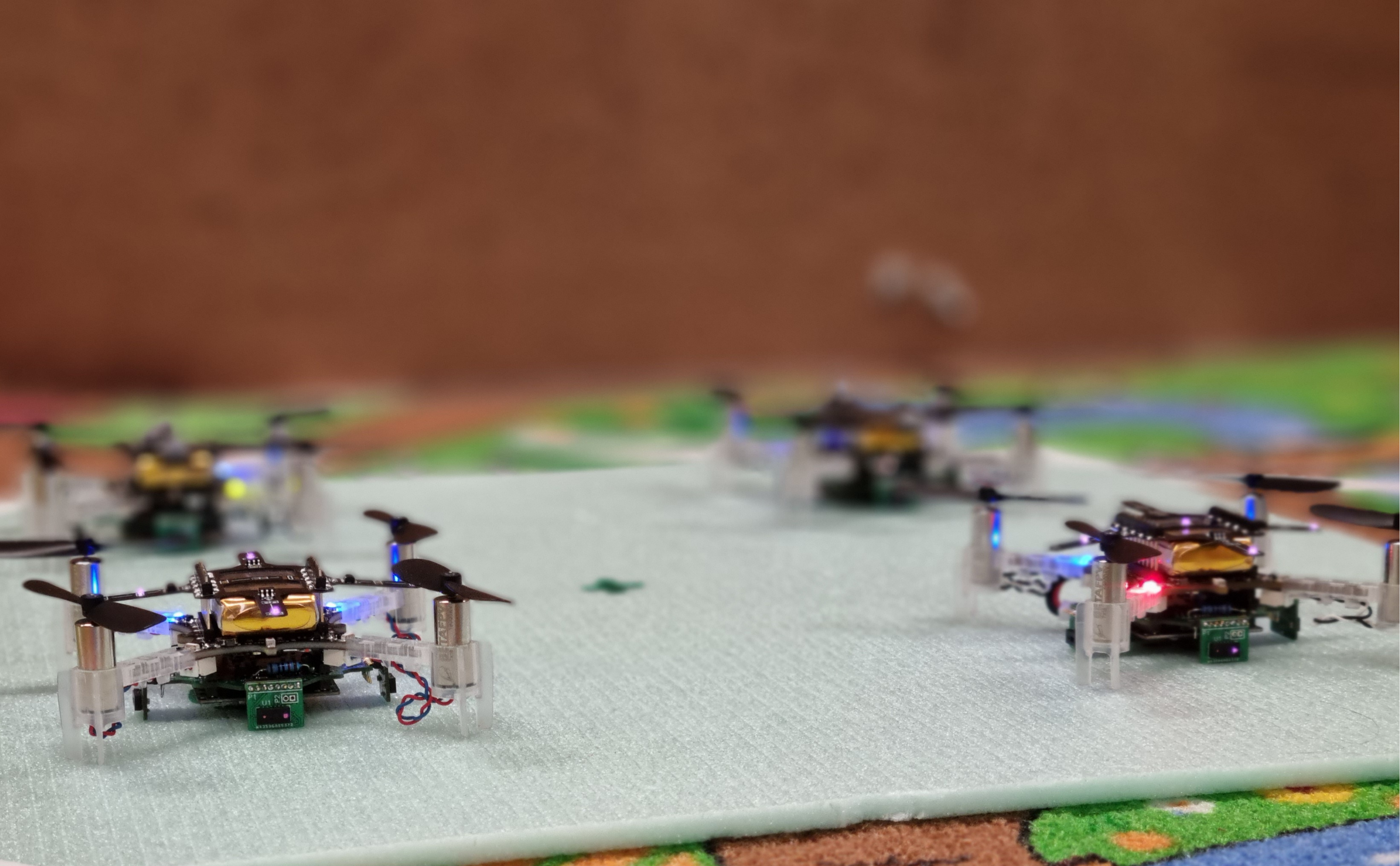}
\par\end{centering}
\caption{The swarm of nano-UAVs before take-off. The platform is based on a commercial Crazyflie 2.1 extended with four VL53L5CX ToF ranging sensors and the flow-deck v2. Total weight at the take-off is \qty{34.8}{\gram}.}
\label{fig:drone_swarm}
\end{figure}

Unmanned Aerial Vehicles (UAVs) have emerged as attractive solutions for several applications that require high maneuverability and scalability, such as distributed inspection and surveillance~\cite{zhang2020drone}. In particular, as defined in ~\cite{wang2020deep}, nano-UAVs have proven to be safe to operate near people due to their reduced weight, i.e., below \SI{50}{\gram}, which makes them an excellent choice for navigating through indoor or cramped environments~\cite{zhou2022efficient}.
Furthermore, they are agile, small enough to fit in the palm of a hand, and their cost-effective hardware facilitates swarm formations. 
An Internet of Robotic Things (IoRT) swarm~\cite{khan2023digital} allows for a decreased latency and a higher probability of reaching the mission objective due to the intrinsic redundancy of having more than one drone~\cite{liu2021cooperative}.
Finding gas leaks~\cite{duisterhof2021sniffy}, localizing survivors in mines~\cite{duisterhof2021tiny}, remote health monitoring of COVID-19 patients~\cite{khan2023digital}, or machine-to-machine cooperation with the Internet of Things (IoT) devices~\cite{polonelli2022open} are only a few examples where nano-UAVs and IoRT~\cite{nawaz2023k}, with a single agent or a swarm, can be employed.
Within such applications, UAVs have to perceive the environment and compute their next movements to enable optimal mission strategy~\cite{loquercio2021learning}. 
However, enabling optimal planning requires good knowledge of the map of the environment as demonstrated in~\cite{loquercio2021learning}.

Simultaneous Localization and Mapping (SLAM) is a class of mapping algorithms that can actively correct for both odometry and mapping errors~\cite{zhou2022efficient}.
In particular, graph-based SLAM is widely used due to its high accuracy and ability to store the whole robot trajectory~\cite{dilshad2022locateuav, bai2021sparse}. 
The algorithm models the drone poses as graph nodes, the odometry measurements as edges, and consists of two main elements: the loop-closure detection, which identifies if the drone has returned to a previously visited location, and the graph optimization, which creates an additional edge as the result of the loop closure, optimizing the previously added poses. Moreover, since mapping requires accurate sensing capabilities that can measure the depth of the environment, Light Detection And Ranging (LiDAR) and stereo cameras~\cite{campos2021orb} are among the most popular approaches that support SLAM on UAVs today. 

Even though the SLAM algorithm paired with LiDARs is widely used~\cite{chang2022lamp,huang2022edge}, it requires computationally intensive and memory-hungry processes, which are not feasible on nano-UAVs due to their limited payload and constrained resources~\cite{niculescu2021improving}.
As an alternative, miniaturized, low-resolution, and energy-efficient Time of Flight (ToF) sensors weighing only \SI{42}{\milli\gram} have been released on the market in the last few years, opening new applications in the field of nano-UAVs~\cite{niculescu2022towards}.
Consequently, recent works have exploited these sensors and enabled SLAM with nano-UAVs. 
However, due to the low sensor resolution (i.e., one depth pixel per sensor), their systems are only capable of mapping simple-geometry environments (e.g., long corridors)~\cite{zhou2022efficient}.
Furthermore, they offload the computation to an external base station via a wireless link, which increases the power consumption and restricts the operating area to the radio range~\cite{zhou2022efficient}. Unlike previous works, we take advantage of the novel VL53L5CX 8$\times$8 ToF matrix sensor, which provides a 64-pixel depth map and offers a Field of View (FoV) of \ang{45}, already characterized in~\cite{niculescu2022towards}. By mounting four such sensors on the nano-UAV (i.e., front, rear, left, right), we increase the FoV and achieve superior loop-closure performance compared to the previous depth-based solutions for nano-UAVs and micro-robots~\cite{zhou2022efficient, karam2022microdrone}.

By projecting the depth measurements acquired over a short time frame into the coordinate system of the drone, we obtain a small map that we call a \textit{scan}. When the drone revisits a location, it acquires another scan and determines the rigid-body transformation with respect to the scan acquired when it first visited that location. This type of loop-closure detection is called \textit{scan-matching}. Furthermore, in this work, we implement and use an optimized version of the Iterative Closest Point algorithm (ICP), a State-of-the-Art (SoA) in scan-matching that works with an arbitrary environment geometry.
In addition, we integrate the ICP with graph-based SLAM, which runs entirely onboard the nano-UAV and achieves full SLAM functionality.

This paper proposes a precise system for distributed indoor mapping with a swarm of nano-UAVs exploiting novel and low-power depth sensors. The entire computation runs onboard the nano-UAVs without relying on any external computer.
Our contributions can be summarized as follows: \begin{enumerate*}[label=(\roman*),,font=\itshape]
\item An optimized implementation and evaluation of the ICP algorithm. It runs entirely onboard a resource-constrained microprocessor in about \SI{500}{\milli\second} for the input size used in our evaluation with an expected translation accuracy of about \SI{3}{\centi\meter}.
Due to the combination with the 8$\times$8 ToF matrix sensors, this is the first work that enables ICP onboard nano-UAVs.

\item A distributed and autonomous exploration policy with obstacle avoidance that allows multiple drones to explore an unknown environment with different flight paths and moving obstacles.

\item The computationally lightweight integration between ICP and SLAM, and its extension to support a swarm of drones. Furthermore, we design and implement a reliable IoRT radio communication protocol that orchestrates how the drones in the swarm exchange poses and scans with each other.

\item An extensive field evaluation that demonstrates the mapping capabilities of our system with a swarm of four nano-UAVs visible in \Cref{fig:drone_swarm}. We prove how our optimized SLAM, combined with ICP, corrects the odometry errors by up to 55\% and aligns the world frames of individual drones to generate coherent maps. To our knowledge, this is the first work that enables onboard mapping with a swarm of nano-UAVs, where robots are working in an IoRT network to collect, analyze, and transmit data.
\end{enumerate*}

%% file: sources/2-related.tex
\section{Related Work}
Standard-size UAVs differ from Micro-Aerial Vehicles (MAVs) and nano-UAVs in terms of size, weight, total power consumption, and  onboard processing capabilities. 
The latter two are directly linked as the budget for  onboard electronics, including sensing and processing, is about $1/10$ of the total motor power~\cite{niculescu2021improving}. Today, most new advancements in the SoA regarding robotic perception and mapping have been demonstrated on standard-size UAVs and MAVs,  which feature a power budget between 50 and \qty{100}{\watt} and a total mass $\geq$\qty{1}{\kilo\gram}, as reported in~\cite{zhou2023racer}. Hence, they feature powerful onboard computing platforms, often equipped with Graphics Processing Units (GPUs) and several gigabytes of memory~\cite{zhou2023racer}. On the other hand, this work focuses on nano-UAVs in particular. They weigh less than \qty{50}{\gram} with a total power budget on the order of 5-\qty{10}{\watt}, of which only \qty{500}{\milli\watt}--\qty{1}{\watt} remain for the sensors and Micro Controller Units (MCUs)~\cite{niculescu2021improving}. Moreover, low-power MCUs mostly support a limited amount of memory, in general between \qty{100}{\kilo\byte} and \qty{500}{\kilo\byte}, a stringent limitation for visual-based perception and mapping. This paper tackles these unsolved research challenges, improving the mapping capabilities of nano-UAVs in single agent or swarm formation.

Previous works on MAVs and UAVs have commonly relied on miniature, conventional \ang{360} LiDAR sensors~\cite{jeong2022parsing} or depth stereo camera~\cite{zhou2023racer} for mapping purposes. In particular, the paper \cite{kumar2017lidar} integrates single-layer LiDAR sensors with inertial measurement units for indoor mapping tasks. The platform used is the commercial DJI Phantom 3 drone with an additional desktop-class Intel i5 processor required onboard. The LiDAR sensor used is \qtyproduct{62 x 62 x 87.5}{\mm} in size and weighs \qty{210}{\g}, while nominally consuming \qty{8.4}{\W} of power. Despite being effective, the setup in~\cite{kumar2017lidar} is unrealistic for a nano-UAV, as depicted in \ref{fig:drone_swarm}, which features a total power budget below \qty{10}{\watt}. Using a similar configuration, the study in \cite{gao20190flying} integrates a multi-layer LiDAR sensor to allow 3D mapping of indoor environments while also relying on a desktop-class Intel i7 processor. Although the LiDAR sensor in this case only consumes \qty{8}{\W} of power, its footprint is larger (\qtyproduct{103.3 x 103.3 x 71.7}{\mm}) and weighs \qty{509}{\g}. On the other hand, authors in \cite{fang2017robust} leverage an RGB-D camera combined with a particle filter for navigating obstructed and visually degraded shipboard environments. The used platform is \qtyproduct{58 x 58 x 32}{\cm} in size, carries more than \qty{500}{\g} of instrumentation, and operates on a high-performance octa-core ARM processor. Again, the payload requirements and computational demands do not fit this paper's scope. SoA mapping strategies are also investigated in the UAV field, as reported in Table~\ref{tab:related}, in terms of sensors, mapping accuracy, swarm size, and power consumed by the computing platforms. In~\cite{causa2022uav}, Causa \emph{et al.} propose a cooperative mapping strategy based on LiDAR and Global Navigation Satellite System (GNSS), relying on a standard size UAV of \qty{3.6}{\kilo\gram} and off-board processing. In \cite{shen2022pgo}, the authors bring the intelligence fully onboard, relying on a power-hungry (i.e., \qty{30}{\watt}) Nvidia Xavier and a VLP-16 LiDAR. Also, in~\cite{huang2022edge}, the mapping algorithm and onboard processing are entrusted to a Jetson TX2 featuring a multi-core Central Processing Unit (CPU) and a GPU, as well as \qty{8}{GB} of memory. Chang \emph{et al.} propose a robust multi-robot SLAM system designed to support robot swarms~\cite{chang2022lamp}; however, results are validated offline on an Intel i7-8750H processor. Although these SoA approaches provide good mapping capabilities in the range of 5 to \qty{20}{\centi\metre}, they involve large and heavy sensors that require processing platforms consuming a few tens of watts. 
These sensors and processors are too power-hungry to operate onboard nano-UAVs and have an unsustainable total weight. Thus, it is clear that standard SoA approaches for autonomous navigation and mapping are not computationally optimized and cannot be deployed in the nano-UAV field~\cite{muller2022robust}, which is the scope of this paper. However, in recent years, lightweight alternative sensing solutions, which are more appropriate for nano-UAV platforms, have become available~\cite{niculescu2022towards, muller2022robust}. Indeed, despite implementing only a basic obstacle avoidance strategy, authors in \cite{muller2022robust} exploit an ultra-lightweight (\qty{42}{\milli\gram}) depth sensor and an optimized navigation algorithm running entirely onboard a nano-UAV. The concept behind \cite{muller2022robust} has been extended for the scope of this paper - enabling distributed mapping on miniaturized robotic platforms.

\begin{table*}[t]
    \centering
    \caption{System and performance comparison between this paper and the State-of-the-Art (SoA) works present in the literature. Onboard processing, sensing elements, mapping accuracy, and system setups are compared.
    \label{tab:related}}
    \begin{tabular}{ >{\centering\arraybackslash}m{1.0cm} >{\centering\arraybackslash}m{2.6cm} >{\centering\arraybackslash}m{2.5cm} >{\centering\arraybackslash}m{1.0cm} >{\centering\arraybackslash}m{1.3cm} >{\centering\arraybackslash}m{1.4cm} >{\centering\arraybackslash}m{1.9cm} >{\centering\arraybackslash}m{1.6cm} >{\centering\arraybackslash}m{1.0cm}} 
        \hline
        \hline
        Work & Onboard processing & Sensor & ToF Pixels & Map accuracy & Field test & Multi-robot Swarm & Power Consumption & System Weight  \\
        \hline
        \multicolumn{9}{c}{Nano-UAV and MAV}\\
        \hline
        \textbf{This work} & \textbf{Yes (Cortex-M4)} & \textbf{$4\times$ ToF VL53L5CX} & \textbf{256} & \textbf{10-15 \qty{}{\centi\meter}} & \textbf{Yes (4 drones)} & \textbf{Yes (up to 20 drones)} & \textbf{240~mW} & \textbf{34.8~g}\\
        \cite{karam2022micro} & No & $4\times$ ToF VL53L1x & 4 & 10-\qty{20}{\centi\meter} & Yes & No & - & \qty{401}{\gram}\\ 
        \cite{duisterhof2021tiny} & Yes (Cortex-M4) & $4\times$ ToF VL53L1x & 4 & - & Yes & No & \qty{240}{\milli\watt} & \qty{31.7}{\gram}\\
        \cite{zhou2022efficient} & No (Intel i7 station) & $4\times$ ToF VL53L1x & 4 & 5-\qty{15}{\centi\meter} & No & No & - & - \\
        \cite{karam2022microdrone} & No & $4\times$ ToF VL53L1x & 4 & \qty{4.7}{\centi\meter} & Yes & No & - & \qty{401}{\gram}\\
        \cite{muller2022robust} & Yes (Cortex-M4) & $1\times$ ToF VL53L5CX & 64 & no map & Yes &No & \qty{320}{\milli\watt} & \qty{35}{\gram} \\
        \cite{niculescu2023nanoslam} & Yes (GAP9) & $4\times$ ToF VL53L5CX & 256 & 8-10 \qty{}{\centi\meter} & Yes & No & \qty{350}{\milli\watt} & \qty{44}{\gram} \\
        \hline
        \multicolumn{9}{c}{Standard-size UAV}\\
        \hline
        \cite{causa2022uav} & No & LIDAR & - & 5-\qty{20}{\centi\meter} & No & Yes & - & \qty{3.6}{\kilo\gram}\\
        \cite{shen2022pgo} & Yes (Xavier) & VLP-16 LiDAR & - & \qty{2.14}{\meter} & Yes & No & \qty{30}{\watt} & $>$\qty{2}{\kilo\gram} \\
        \cite{huang2022edge} & Yes (Jetson TX2) & RP-LIDAR & - & - & Yes & Yes up to 50 & \qty{7.5}{\watt} & \qty{1.8}{\kilo\gram} \\
        \cite{chang2022lamp} & No (Intel i7 station) & LIDAR & - & 15-\qty{20}{\centi\meter} & No & Yes & - & -\\
        \cite{zhou2023racer} & Yes (Jetson TX2) & Intel RealSense D435 & - & - & Yes & Yes & \qty{7.5}{\watt} & \qty{1.3}{\kilo\gram} \\

        \hline
        \hline
    \end{tabular}
\end{table*}

In~\cite{karam2022micro}, simple ToF ranging sensors facing forward, backward, left, and right on a Crazyflie drone have been used to implement mapping. Although this work implements basic obstacle avoidance, manual piloting from a ground station is required, and no method for compensating odometry drift is implemented. Using the same hardware configuration, authors in \cite{duisterhof2021tiny} implement obstacle avoidance for source seeking using deep reinforcement learning. However, since this method does not map the environment, the path is often suboptimal. Mapping using these simple ranging sensors is possible, but the low resolution of the acquired data means that longer flight times are required to approach the mapping fidelity of traditional drone-based systems while still suffering from odometry drift. Mapping methods can be improved by applying SLAM algorithms to correct pose estimation errors, which is the main scope of this paper.

A mapping pipeline typically requires two stages: first, the robot must close the loop using scan-matching and then use the scan-matching information to correct the trajectory. ICP represents the SoA for performing scan-matching, and multiple variations were proposed in the literature. The authors in \cite{pan2021mulls} propose MULLS ICP, which formulates the scan-matching as a weighted least squares problem. Although it provides accuracy in the order of a few tens of centimeters, it requires tuning several parameters, and thus, it was discarded for this work.
The authors in \cite{dellenbach2022ct} propose CT-ICP, a particular form of ICP that also corrects the measurement distortion introduced by robot motion. They achieve a mean translation error of \qty{0.55}{\meter}, but their approach requires knowing the robot motion profile. Furthermore, they determine the correspondences between the scan points using the point-to-plane approach, which is computationally demanding for a nano-UAV onboard MCU. Recently, the authors in \cite{vizzo2023kissicp} released Kiss-ICP, which overcomes the issues associated with the previous ICP works and achieves a translation error of \qty{0.49}{\meter}. Furthermore, they use the point-to-point correspondence between the scan points, which is computationally faster than the point-to-plane or point-to-line alternatives. Therefore, this approach was employed for the scope of this paper. All the mentioned scan-matching works use ICP paired with LiDARs to enable odometry estimation while the robot is moving. In our work, we use ICP paired with the same correspondence mechanism as in \cite{vizzo2023kissicp}, but only perform scan matching when a loop closure is detected. 
Furthermore, within this paper scenario, the drone is stationary while acquiring the scans, leading to a translation error of less than \qty{10}{\centi\meter} despite the sparse input information provided by the ToF sensors.
The execution time of the SoA is in the range of 26-\qty{83}{\milli\second}~\cite{dellenbach2022ct, vizzo2023kissicp, pan2021mulls}, when running on commodity computers with general-purpose CPUs, such as the Intel i7-7700HQ. On the other hand, our system requires an execution time of about \qty{500}{\milli\second} with a power consumption that is two orders of magnitude smaller with respect to an Intel i7 processor.

The sub-field of SLAM problems known as "SLAM with sparse sensing"~\cite{matsuki2021codemapping} explores more challenging applications where robots receive data points with a lower frequency and accuracy, as is the case for sensing solutions suitable for nano-UAVs~\cite{sonugur2022review}.  For instance, the work in \cite{beevers2006slam} proposes to solve this problem by extracting line features as landmarks and applying a Rao-Blackwellized Particle Filter (RBPF)~\cite{doucet200rao}. However, particle filters are not a favorable approach for larger maps since large numbers of particles are required to maintain accuracy~\cite{wilbers2019comparison}. With this in mind, results in \cite{zhou2022efficient} demonstrate the efficient offline use of graph-based SLAM on sparse data with a novel front-end for scan matching. Although this method uses an Intel i7 desktop-class processor, it was also evaluated using data collected from a Crazyflie drone using single-zone ToF sensors, a setup similar to this paper. Following up on~\cite{karam2022micro}, the method presented in~\cite{karam2022microdrone} shows a similar approach to applying SLAM to compensate for odometry drift; however, it is still being computed offline. The approach is further limited by the single-zone sensors, which are mounted on a larger drone (\qtyproduct{182 x 158 x 56}{\milli\metre}) based on the Crazyflie Bolt.

Works such as~\cite{chengbo2019research} used ranging sensors to implement SLAM  onboard differential wheeled robot platforms using embedded application processors. Furthermore, authors in~\cite{zhu2021visual} use an application processor and cameras to implement visual SLAM onboard a wheeled consumer robot platform, while in~\cite{du2022integrated} a visual SLAM algorithm optimization is discussed to exploit better the resources of an Open Multimedia Application Platform (OMAP) processor. On the other hand, \cite{beevers2008embedded} shows an MCU-based solution using a particle filter-based approach~\cite{beevers2006slam} that relies on extracting line features. Similarly, a different method called orthogonal SLAM is proposed in~\cite{alpen2010onboard} and executed on an MCU onboard a drone in~\cite{alpen2013autonomous}. However, this method assumes that the extracted line features are always orthogonal. In contrast, our approach makes use of a fully onboard, highly optimized implementation of graph-based SLAM~\cite{grisetti2010tutorial} with scan matching using ICP~\cite{vizzo2023kissicp} adapted for a novel sensing solution on a \qty{34.8}{\gram} nano-UAV.  

Although visual and sparse-distance SLAM mapping is a well-known topic in the robotics community, the computational load and the related processing latency are still a concern in the UAV field~\cite{loquercio2021learning}. Moreover, the current research trend pushes for collaborative mapping among a group of agents~\cite{jinqiang2021self}, also referred to as IoRT~\cite{khan2023digital}, often composed of a heterogeneous set of flying robots and sensing elements~\cite{zhou2023racer}. As shown by Table~\ref{tab:related}, only a few works in the literature propose a distributed mapping solution based on a UAV swarm that is successfully implemented in a real experiment~\cite{huang2022edge}, in particular without relying on any external infrastructure~\cite{chang2022lamp,zhou2023racer}. In the nano-UAV field, lightweight methodologies and field tests demonstrating mapping capabilities are scarce~\cite{duisterhof2021tiny,zhou2022efficient}. A recent publication implementing onboard SLAM with nano-UAVs is titled NanoSLAM~\cite{niculescu2023nanoslam}, which reaches \qty{10}{\cm} mapping accuracy. However, despite the relevant results regarding system integration and latency, NanoSLAM supports only one agent and necessitates a supplementary co-processor, which increases the total weight at take-off to \qty{44}{\gram} as shown in \Cref{tab:related}, thus decreasing the flight time concerning the solution proposed in this paper. Moreover, \cite{niculescu2023nanoslam} shows how the mapping coverage is directly proportional to the nano-UAV battery lifetime, flying at relatively low speed (below \qty{2}{m/s}). Therefore, since the flight time is bounded by the technology limitation of the battery capacity and the frame weight, the alternative to increase the mapping coverage is to parallelize the task over a swarm of nano-UAVs. To the best of our knowledge, no works investigating a mapping system based on a swarm of nano-UAVs are present in the literature. Thus, this paper proposes the first study, implementation, and field results enabling fully onboard and distributed mapping for the nano-UAV ecosystem. We demonstrate the possibility of supporting up to 20 agents relying only on a \qty{34.8}{\gram} hardware platform for sensing, processing, and communication. The achieved accuracy is aligned with the SoA for MAVs and standard-size UAVs, with an average error on the order of \qty{12}{\cm}. The proposed system, including the lightweight perception framework implemented in this paper, paves the way for enhancing the autonomous capabilities of nano-UAVs by improving optimal path planning and IoRT multi-agent collaboration. 

%% file: sources/3-background.tex
\section{System setup} \label{sec:background}

This section describes the nano-UAV platform selected for this work, deepened in Section~\ref{sec:crazyflie}, and the supplementary commercial and custom sensors, in Section~\ref{sec:deck}, used as the basis for our experiments. The onboard computational and sensing capabilities are discussed, as well as the hardware's total weight.
\subsection{Crazyflie Nano-UAV}
\label{sec:crazyflie}
The Crazyflie 2.1 is an open nano-UAV platform from Bitcraze commonly used in research. Its mainboard also acts as the airframe and includes an Inertial Measurement Unit (IMU), a barometer, a radio (Nordic nRF51822), and an STM32F405 MCU (168 MHz, 196 kB RAM) that handles sensor readout, state estimation, and real-time control. The drone features extension headers that can be used to add commercially available decks (i.e., plug-in boards) to improve its sensing capabilities. In this work, the drone was equipped with the commercial Flow-deck v2, featuring a downward-facing optical flow camera (i.e., PMW3901)  and single-zone ToF ranging sensor (i.e., VL53L1CX), which improve velocity and height measurements through sensor fusion by the onboard Extended Kalman Filter (EKF).
The drone's MCU and the mentioned sensors consume about \qty{100}{\milli\watt}.
The Crazyflie was further equipped with a custom deck designed and presented in this work featuring four VL53L5CX ToF ranging sensors -- each sensor consuming an additional power of \qty{100}{\milli\watt}. The total weight at take-off is \qty{34.8}{\g}, including all the hardware used for the scope of this paper. The final system setup is depicted in \Cref{fig:drone_swarm}. With a \qty{350}{\milli\ampere\hour} battery, the drone can fly for about \qty{6}{\minute}.

\subsection{Custom Quad ToF Deck}
\label{sec:deck}
The VL53L5CX is a multi-zone 64-pixel ToF sensor featuring an extremely lightweight package, with a total weight of \qty{42}{\milli\gram} and an accuracy of \qty{4}{\centi\meter}~\cite{muller2022robust}.
Its performance in the field of nano-UAVs was characterized in~\cite{niculescu2022towards}. 
The maximum ranging frequency for the full resolution of 8$\times$8 pixel is \qty{15}{\Hz}, and the Field of View (FoV) is \ang{45}. 
Moreover, the VL53L5CX provides a pixel validity matrix paired with the 64-pixel measurement matrix, which automatically flags noisy or out-of-range measurements. 
To enable the use of the multi-zone ranging sensors with the Crazyflie, we have created a custom deck designed specifically for the VL53L5CX ToF sensor. 
It can be used simultaneously with the Flow-deck v2 and incorporates four VL53L5CX sensors that face forward, back, left, and right, allowing for the detection of obstacles in all directions. 
Although lower than in conventional LIDARs, the 8$\times$8 resolution is enough to enable accurate scan-matching. 
Unlike LIDARs, the novel multi-zone ToF sensor is a perfect trade-off between resolution, accuracy, weight, and power consumption for use onboard nano-UAVs. Moreover, we incorporate a \qty{4}{MB} Serial Peripheral Interface (SPI) flash storage device to store the acquired measurements. The memory type was chosen based on the interfaces available onboard the drone (i.e., SPI). Among the available memories with the SPI interface, the flash memory typically has more capacity for the same size than the SRAM and a much smaller latency than a microSD card. The final design of the custom ToF deck only weighs \qty{5.1}{\g}.

%% file: sources/4-algorithms.tex
\section{Algorithms} \label{sec:algorithms}

This section provides the theoretical background of the lightweight algorithms we implement onboard and how they fit together to solve the mapping problem.
Generating an accurate map requires good accuracy of the drone's position estimate, which can be impacted by odometry drift, leading to warped maps. We show how we leverage and combine the information from multiple ToF multi-zone sensors to achieve scan matching using the ICP algorithm. While ICP can correct the odometry errors in the re-visited locations, we show how it can be combined with SLAM to correct the whole past trajectory and therefore, enable accurate mapping. Moreover, ICP is also used to combine data collected from different nano-UAVs in the swarm.

\subsection{Onboard State Estimation} \label{sec:state-estimation}
\begin{figure} [t]
\begin{centering}
\includegraphics[width=\columnwidth]{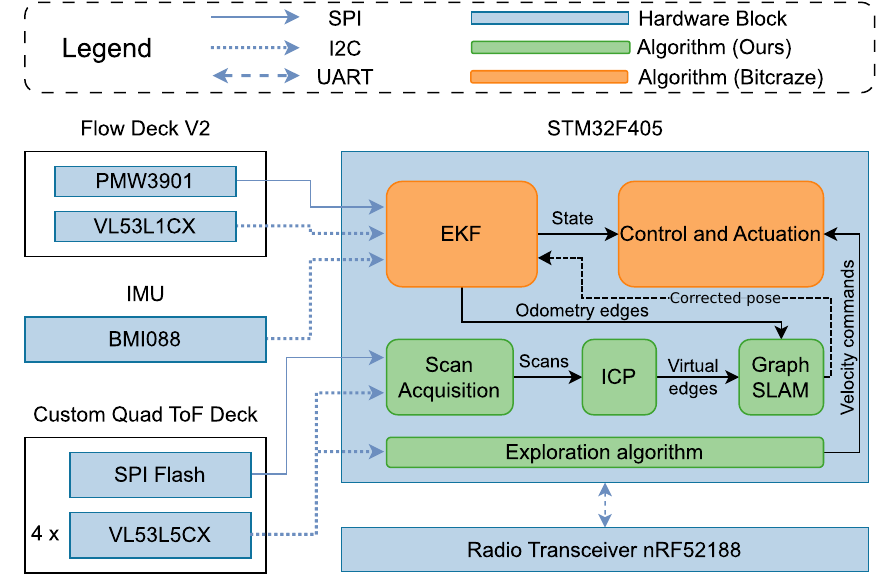}
\par\end{centering}
\centering{}
\caption{Overview of the system architecture, including a hierarchical description of the algorithmic pipeline and data interconnections.}
\label{fig:state-estimation}
\end{figure}

The state estimation relies entirely on onboard sensors, leveraging the IMU, the optical flow camera, and the downward-pointing single-pixel ToF sensor.
The IMU provides 3D acceleration and angular velocity, which are crucial for attitude estimation.
The downward-pointing ToF sensor provides absolute height measurements, and its combination with the optical-flow camera enables body-frame velocity measurements.
All the sensor information is fused by the onboard EKF, which provides complete state estimation -- i.e., position, velocity, and orientation.
The EKF and the state controller are functionalities implemented in the open-source Crazyflie firmware.
Figure~\ref{fig:state-estimation} shows an overview of the system architecture, where the orange software blocks belong to the base firmware.
The blocks illustrated in green represent our additions, implemented as separate FreeRTOS tasks and discussed in this section.

\subsection{Scan Frames and Scans} \label{sec:algo-scans}

In this work, we tackle the mapping problem in 2D, and therefore, the localization and mapping are performed in one plane. Since each ToF sensor provides an 8$\times$8 distance matrix, we need to reduce it to an 8-element array that is compatible with our 2D scenario. Hence, we derive a single measurement from each column of the ToF matrix. To this end, we only consider the center four pixels from each column, discard any invalid pixels, and take the median of the remaining values. Should none of the pixels be valid, the entire column is discarded. This occurs, for example, when there is no obstacle within the \qty{4}{\m} range of the sensor. The mechanism of selecting the median pixel from each row provides a low-pass filtering effect, which mitigates the effect of the high-frequency noise in the nano-UAV's motion.
Furthermore, each sensor is polled before the mission starts to ensure the proper functionality and avoid possible installation issues.

\begin{figure*}[t]
    \centering
    \begin{tabular}{cc}
\begin{subfigure}{0.45\textwidth}
    \centering
    \smallskip
    \includegraphics[width=\linewidth]{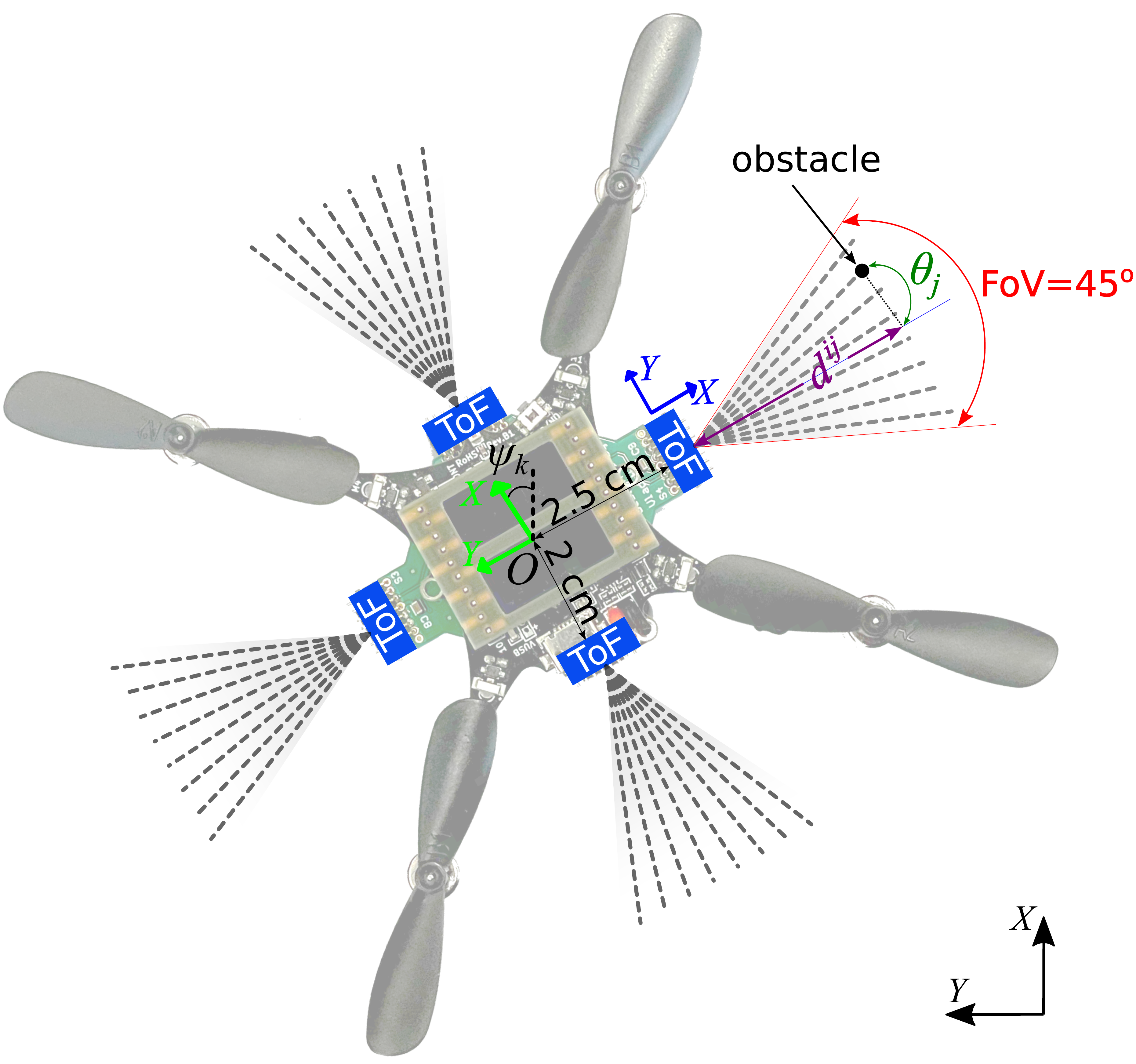}
    \caption{Representation of the four onboard ToF sensors and of the variables in Equation~\ref{eq:scan-proj}.}
    \label{fig:scan-frame}
\end{subfigure}
    &
        \begin{tabular}[b]{c}
        \smallskip
            \begin{subfigure}[t]{0.45\textwidth}
                \centering
                \includegraphics[width=\textwidth]{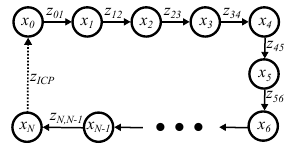}
                \caption{Pose graph for a single drone.}
                \label{fig:single-agent}
            \end{subfigure}\\
            \begin{subfigure}[t]{0.45\textwidth}
                \centering
                \includegraphics[width=\textwidth]{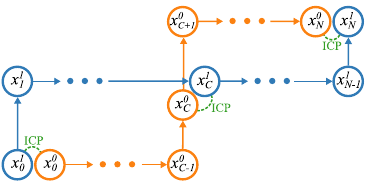}
                \caption{Pose graph for multiple drones.}
                \label{fig:multi-agent}
            \end{subfigure}
        \end{tabular}\\
    \end{tabular}
    \caption{Illustrations of the four ToF sensors (Figure~\ref{fig:scan-frame}) and the pose graphs (Figures~\ref{fig:single-agent} and \ref{fig:multi-agent}).}
\end{figure*}

Let $\vec{x}_k=(x_k,y_k,\psi_k)$ be the state of the drone (i.e., \textit{pose}) at timestamp k expressed in the world coordinate frame -- in practice, the state information is provided by the onboard EKF. Furthermore, let $i\in \{1,2,3,4\}$ describe index among sensors and $j\in \{1,2, \ldots 8\}$ the index among the zones of each sensor. Thus, $d_k^{ij}$ represents the distance the $i$-th sensor provides for the $j$-th zone.
Equation~\ref{eq:scan-proj} provides the function that projects a distance measurement $d_k^{ij}$ acquired at pose $\vec{x}_k$ into the world coordinate frame.
Figure~\ref{fig:scan-frame} provides a graphical representation of the variables in Equation~\ref{eq:scan-proj}, and the world and the drone's body frames are represented in black and green, respectively.
The heading angle $\psi_k$ represents the rotation between the drone's body frame and the world's frame, $\beta_i \in \{ \SI{0}{\degree},  \SI{90}{\degree}, \SI{180}{\degree}, \SI{270}{\degree}\}$ represents the fixed rotation of each sensor with respect to the drone's body frame and $\mat{R}$ is the 2D rotation matrix. Furthermore, $o_x^i$ and $o_y^i$ represent the offset of each ToF sensor $i$ with respect to the drone's center $O$ expressed in the body frame of the drone.
$d_{ij}$ is the projection of a distance measurement on the $OX$ axis of the ToF sensor's coordinate frame (marked with blue in Figure~\ref{fig:scan-frame}), and the sensor directly provides it. The y-coordinate of the measurement in the same coordinate frame is calculated as $tan(\theta_j) \cdot d^{ij}$, where $\theta_j$ is the angle of the zone.
\begin{equation} \label{eq:scan-proj}
\vec{s}(\vec{x}_k, d_k^{ij})
= 
\begin{pmatrix}
x_{k}\\
y_{k}
\end{pmatrix}
+
\mat{R}_{(\psi_k + \beta_i)}
\begin{pmatrix}
d_k^{ij} + o_{x}^i \\
tan(\theta_j) \cdot d_k^{ij} + o_{y}^i
\end{pmatrix}
\end{equation}

We define the \textit{scan frame} as the set containing the 2D projection of each distance measurement for a particular timestamp $k$.
The size of a scan frame is, at most, 32 points -- 4 sensors $\times$ 8 zones -- as some distance measurements might be invalid and, therefore, not included in the scan frame. A scan frame is therefore obtained by applying Equation~\ref{eq:scan-proj} for each of the 32 points. The projection of the distance measurements is required for the scan-matching algorithm. However, the 32 points in a scan frame are still too sparse to compute accurate scan matching. We solve this problem by stacking 15 scan frames in a set that we call a \textit{scan}. Specifically, once the nano-UAV decides to acquire a scan, it appends a new scan frame with a frequency of \qty{7.5}{\hertz} until it reaches the count of 15. To avoid synchronization problems, the four ToF sensors start data acquisition simultaneously. During the mission, the time misalignment between the sensors is about \qty{9}{\milli\second}, which would result in sub-centimeter errors given the low speed of nano-UAVs. Furthermore, to mitigate these effects even more, we command the drone to stop while acquiring a scan. 

The four ToF sensors have a cumulative FoV of \ang{180} -- \ang{45} per sensor.
To maximize the scan coverage, the drone also rotates in place by \ang{45} on the yaw axis while acquiring the scan, resulting in a \ang{360} coverage.
We empirically determined that the odometry errors introduced by spinning the drone during scan acquisition are negligible, leading to an increase of about \qty{2}{\centi\meter} and \qty{1.5}{\degree} for the positioning and heading errors, respectively.
In conclusion, we can define a scan as $\mat{S}_k=\{\vec{s}(\vec{x}_{\tilde k}, d_{\tilde k}^{ij}) \mid i \leq 4; j \leq 8;  k \leq \tilde k < k+15; i,j,\tilde k \in \mathbb{N}^* \}$ and each scan $\mat{S}_k$ has an associated pose $\vec{x}_k$, acquired right before the scan acquisition starts.
The resulting size of a scan is, at most, 480 points. This setting was empirically selected based on the trade-off between the need to have sufficient points for ICP-based scan-matching to produce accurate solutions and the memory footprint.

\subsection{Iterative Closest Point} \label{sec:icp}
Scan matching is the process of aligning two scans, such that overlapping features line up correctly. 
The result of scan matching is the optimal rotation and translation that is applied to one scan, such that its features overlap with the other scan.
Since each scan has an associated pose, the transformation between two scans is the same as the transformation between their associated poses.
This is critical for correcting errors in odometry estimations that lead to the misalignment of poses that were acquired at different times or locations. 
We choose the ICP algorithm for scan matching and describe the theoretical study of ICP optimized for a practical implementation onboard nano-UAVs.

We define two scans \(P = \{ \boldsymbol{p}_1, \ldots, \boldsymbol{p}_N\}\) and \(Q = \{ \boldsymbol{q}_N, \ldots, \boldsymbol{q}_M\}\) and omit the time index in the notation for simplicity. Scan matching can be formulated as a least squares problem, and finding the optimal transformation between $P$ and $Q$ is equivalent to solving the optimization problem in Equation~\ref{eq:icp-obj}~\cite{vizzo2023kissicp}. However, this equation assumes known data associations -- i.e., which point in scan Q corresponds to each point in scan P. 
Under this assumption, solving Equation~\ref{eq:icp-obj} leads to the optimal solution without requiring an initial guess.
In the ideal case, when the scans are identical, rotating each scan point $Q$ using $\mat{R}^*$ and then adding the translation factor $\vec{t}^*$ should result in a perfect overlap with scan $P$ and the same holds for the poses $\vec{x}_Q$ and $\vec{x}_P$ that are associated with the two scans.
\begin{equation} \label{eq:icp-obj} 
    \mat{R}^*,\vec{t}^* = \argmin_{\mat{R},\vec{t}} \sum{\lVert \vec{q}_i - (\mat{R} \vec{p}_i + \vec{t}) \rVert^2}
\end{equation}

However, no prior knowledge of correspondences is available in real-world applications, and a heuristic is required. A common method for establishing the correspondences for a given point in a scan is to find the closest point in the other scan in terms of Euclidean distance~\cite{vizzo2023kissicp}. In practice, this step is performed using a double loop, and it, therefore, comes with a quadratic complexity. Furthermore, it typically accounts for more than 90\% of the total ICP execution time. Once the correspondences are calculated, Equation~\ref{eq:icp-obj} determines the optimal transformation. The iterative process between calculating the correspondences and computing the optimal transformation, until the mean Euclidean distances between the correspondence pairs reach the minimum, is called ICP. Its accuracy is mainly impacted by how precisely the correspondences are determined.

\subsection{Simultaneous Localization and Mapping}
Producing an accurate map requires precise robot position estimation.
In most indoor scenarios (including ours), the position and heading of the drone are computed by integrating velocity and angular velocity measurements, respectively -- typically performed by the onboard EKF.
However, these measurements are affected by sensor noise, and integrating noisy data over time results in drift. We employ SLAM to correct the errors in the trajectory of the robot caused by imprecise odometry measurements. However, this approach is also valid in the case of absolute range-based positioning systems~\cite{polonelli2022open}.

We employ the algorithmic approach to graph-based SLAM introduced in \cite{grisetti2010tutorial} and implement an optimized version that allows the algorithm to run onboard resource-constrained nano-UAVs. Within this approach, the UAV's poses $\mat{X} = \{ \boldsymbol{x}_1, \ldots, \boldsymbol{x}_N\}$ are modeled as graph nodes, and the odometry measurements $\boldsymbol{z}_{ij}$ as graph edges. An odometry measurement $\boldsymbol{z}_{ij}=(\Delta_x,\Delta_y,\Delta_{\psi})$ is expressed in the coordinate frame of $\boldsymbol{x}_{i}$ and represents the relative transformation between poses $\boldsymbol{x}_{i}$ an $\boldsymbol{x}_{j}$.
The work in \cite{grisetti2010tutorial} formulates the SLAM as a least-squares optimization problem that determines the optimal poses given the pose $\boldsymbol{x}_{0}$ and a set of edges $\boldsymbol{z}_{ij}$.
The optimization problem is given by Equation~\ref{eq:slam-opt}, where the number of terms in the sum is equal to the number of edges in the directed graph.
$\Omega$ is the diagonal information matrix, which weighs the importance of each term in the sum.
\begin{align} \label{eq:slam-opt}
\vec{e}_{ij} &= \vec{z}_{ij} - \vec{\hat z}_{ij}(\vec{x}_{i}, \vec{x}_{j}) \nonumber \\
\mat{X}^* &= \argmin_{\mat{X}} \sum_{i,j} \vec{e}_{ij}^T \Omega \vec{e}_{ij} 
\end{align}

Figure~\ref{fig:single-agent} shows a simple pose graph example, but representative of any real-world scenario.
Since any node is connected to the previous one by an odometry measurement, at least one path exists between any two given nodes.
The optimization solution given by \cite{grisetti2010tutorial} to the problem in \ref{eq:slam-opt} is based on the Gauss-Newton iterative method, and an initial guess is required for the poses, which we obtain from the nano-UAV's odometry measurements. Since the drone directly obtains the odometry measurements $\vec{z}_{01}, \vec{z}_{12}, \ldots \vec{z}_{N-1, N}$ from the state estimator,  it is straightforward to compute the initial guess of the poses $\vec{x}_{1}, \ldots \vec{x}_{N}$ by forward integration with respect to $\vec{x}_{0}$.
This is the best guess that we have, which is quite close to the real values assuming that the odometry drift is in bounds -- typically below \qty{0.5}{\meter}. So far, solving the optimization problem in Equation~\ref{eq:slam-opt} for the measurements $\vec{z}_{ij}$ would lead to no change in the poses because the pose values after the forward integration are already in agreement with the measurements. Assuming that poses $\vec{x}_{N}$ and $\vec{x}_{0}$ are close enough in terms of Euclidean distance, one can use the ICP algorithm introduced in Section~\ref{sec:icp} to generate a direct relation between the two poses -- which we call a \textit{virtual edge} or a \textit{constraint}.
The virtual edges take part in the optimization process just as the other edges, as additional terms to the sum in Equation~\ref{eq:slam-opt}.
However, since the scans are very accurate, so is the result of the ICP, and, therefore, the virtual edge measurements are more precise than the edges associated with odometry measurements. 
Consequently, the information matrix $\Omega$ associated with the virtual edges takes the value $20\mat{I}$, while the information matrix for the odometry edges is $\mat{I}$. These values are determined empirically. Figure~\ref{fig:state-estimation} shows the interaction between the EKF, ICP, and SLAM.
The EKF provides the raw poses as input to the SLAM algorithm, which are used to derive the odometry edges. Furthermore, by acquiring scans in revisited places, the ICP block derives relative constraints between poses, which are also communicated to the SLAM block as virtual edges. The SLAM block optimizes the poses using the odometry and virtual edges and provides a new set of optimized poses.
Optionally, the most recent optimized pose can be used to correct the accumulated odometry drift by overwriting the current estimate of the EKF.

So far, we have presented how we integrate the graph-based algorithm presented in~\cite{grisetti2010tutorial} with ICP to optimize a single drone agent's trajectory (i.e., poses). In the following, we show how we apply SLAM when using multiple drones. Figure~\ref{fig:multi-agent} illustrates the pose graphs of two drones following two independent trajectories. For the sake of readability, we omit the notation for the edge measurements and use the superscript in the pose notation to indicate which drone the pose is associated with. Within the SLAM formulation introduced so far, the two trajectories would result in two disconnected graphs. However, knowing the locations where the trajectories intersect, the drones can use ICP to determine how closely located poses (e.g., $\vec{x}_0^0$ and $\vec{x}_0^1$) relate and therefore create a connection between the two graphs for each intersection point. This approach enables both loop closure and alignment of the trajectories of multiple drones in a common coordinate system. While the scenario in Figure~\ref{fig:multi-agent} only refers to two drones, the approach scales for any number of drones in the swarm. Putting together the pose-graphs associated with multiple drones results in a larger graph that we call \textit{the global graph}. 

\begin{figure*} [t]
\begin{centering}
\includegraphics[width=\textwidth]{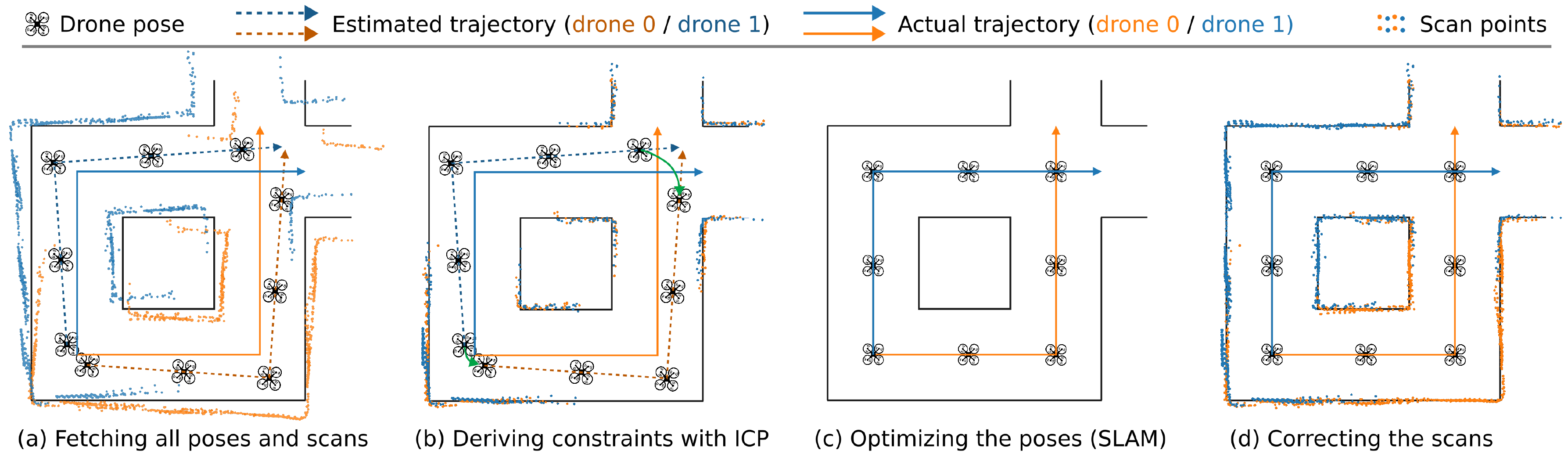}
\par\end{centering}
\caption{The steps involved in the process of collecting and optimizing the poses and scans to generate a coherent map. This example uses two drones, and their estimated and actual trajectories are represented with dashed and solid lines, respectively. The walls of the environment are illustrated with black solid lines, while the ICP constraints are shown as green arrows.}
\label{fig:mapping-steps}
\end{figure*}

Figure~\ref{fig:mapping-steps} illustrates a simplified example of how the information from multiple drones is merged to optimize each drone's trajectory and generate a collective map.
The map is constructed by using the scans acquired by all drones and applying the SLAM optimization computed on one drone, which we call \textit{the main drone}.
Thus, in the first step the main drone -- \textit{drone 0} in this example -- collects the scans and poses from \textit{drone 1}, which together constitute the raw data, as shown in Figure~\ref{fig:mapping-steps}-(a).
We recall that every pose has an associated scan because whenever the drone logs a new pose, it also spins by \qty{45}{\degree} to obtain the scan.
However, the scan is not directly acquired, but computed by projecting the ToF measurements in the world frame using Equation~\ref{eq:scan-proj} -- this is performed in a distributed fashion, to reduce the computational load of the main drone.

Consequently, the data points in Figure~\ref{fig:mapping-steps}-(a) are obtained by merging the five scans from drone 0 (in orange) with the five scans obtained from drone 1 (in blue).
Due to odometry drift, the maps provided by each drone are not aligned.
Next, the main drone derives virtual edges between the overlapping scans using ICP -- represented with green arrows in Figure~\ref{fig:mapping-steps}-(b).
In this example, the overlapped scans belong to different drones, but as shown earlier in this section they could also belong to the same drone when a location is revisited.
Figure~\ref{fig:mapping-steps}-(b) shows only the overlapped scan pairs, acquired in the bottom left and top right corners.
The base-poses and virtual edges represent the prerequisites for optimizing the pose-graph using SLAM, which leads to the optimized set of poses that are shown in Figure~\ref{fig:mapping-steps}-(c).
The same corrections applied to each pose can subsequently be applied to the corresponding scans, facilitating the creation of a coherent global map as shown in Figure~\ref{fig:mapping-steps}-(d).

Since the size of the resulting global graph represents the sum of all subgraph sizes, there is no difference in computation and memory requirements between a swarm of m drones with N poses each and a single drone with m$\cdot$N poses. Since the graph optimization is only executed on a single drone at a time, the main limitation of the system is the maximum number of poses that the SLAM algorithm can optimize within an acceptable time. However, this limitation could be further relaxed by dividing the global graph. For example, in the situation depicted in Figure~\ref{fig:multi-agent}, optimizing the graph $\vec{x}_0^{1,0}, \vec{x}_1^{1,0}, \ldots \vec{x}_C^{1,0}$ and then using the optimized value of $\vec{x}_C^{1,0}$ as a constraint for optimizing $\vec{x}_C^{1,0}, \vec{x}_{C+1}^{1,0}, \ldots \vec{x}_N^{1,0}$ would result in very similar results to the case of optimizing the whole global graph at once.

\subsection{Autonomous Exploration} \label{sec:autonomous-exploration}
Given that the environment is unknown, an autonomous exploration policy is needed to guide the nano-UAVs through the unfamiliar environment. 
The policy is deliberately designed to choose different paths for each drone, assuming sufficient distinct paths exist. 
To simplify the evaluation, we assume an environment consisting of perpendicular walls and equal-width corridors.
For this purpose, we develop a simple exploration policy that can drive the drone along the $x$ or $y$ axis.
While it can still be used in environments with non-perpendicular walls, this would lead to suboptimal ``zig-zag'' trajectories.
However, minor adjustments could enable the exploration strategy to cope with arbitrary geometry environments if desired. The autonomous exploration policy is based on following walls or corridors and relies on the ToF sensor information to avoid obstacles. More in detail, it computes the minimum distance from the matrix in each direction and uses this information to make decisions.
The motion commands are separated along two axes: primary and secondary. 
The primary axis is the direction the drone aims to explore. 
The secondary axis is perpendicular, and motion along this axis should be minimal. 

The motion along the \textit{secondary axis} is commanded so that the drone maintains a constant distance from the walls. 
A proportional velocity controller with a of gain $v_{stab} = \qty{2}{\per\s}$ determines the velocity set point based on the distance to the walls on either side -- i.e., $d_L$ and $d_R$.
When the drone detects that it is located in a corridor (i.e., walls on both sides), it attempts to center itself by targeting a velocity $v_{sec} = d_L - 0.5\cdot(d_L + d_R)) \cdot v_{stab}$. 
Otherwise, if there is a wall within \SI{1}{\meter} of either side of the drone, it attempts to hold a target distance of $d_{wall} = \SI{0.5}{\meter}$ to the wall, by applying $v_{sec} = (d_{wall} - d_{L,R}) \cdot v_{stab}$. 
These values are chosen assuming that the width of corridors is approximately \SI{1}{\meter}, but the parameters can be adapted for different environments.
The wall-following is effective for scan matching since walls and corners typically provide feature-rich scans while avoiding frequent out-of-range measurements associated with large open spaces. 


The \textit{primary axis} describes the direction of exploration. Its control is based on waypoints. 
The navigation policy continuously seeks the next waypoint, which is always located about \SI{1}{\meter} away from the previous one in the direction of motion.
In each waypoint, the drone adds a new pose to its graph and acquires and stores a new scan.
Equation~\ref{eq:v_pri} describes the proportional velocity control along the primary axis, where $v_{exp}$ is the nominal exploration velocity, $d_w$ is the remaining distance to the next waypoint and $d_{slow} = \qty{0.75}{\m}$ determines the distance to a waypoint where the drone starts slowing down. 
Additionally, increases in velocity are limited to an acceleration of $a_{exp} = \qty[per-mode=symbol]{0.5}{\m\per\square\s}$, to smooth the drone's motion.
When the drone faces an obstacle in the direction of exploration before reaching the next waypoint, it immediately chooses a new perpendicular direction for exploration. The primary and secondary axes are switched, and exploration continues. If the only option is to go back the way the drone came (i.e., dead-end), it lands and finishes exploring. The choice of the new direction depends on the physical environment and a predefined preference that is set differently across individual drones in order to prevent multiple drones from exploring the same path.
\begin{equation} \label{eq:v_pri}
    v_{pri}(d_w)
    =
    \begin{cases}
        (\frac{d_w}{d_{slow}} + 0.1) v_{exp} & \quad\text{if } d_w < d_{slow} \\
        v_{exp} & \quad\text{otherwise} \\
    \end{cases}
\end{equation}

Given that we are interested in operating a swarm of drones, and in most situations, they start from the same location, it would be sub-optimal if all drones had identical flight paths. 
At the same time, for testing purposes, individual drones' trajectories should be predictable and repeatable. In addition to alternating the steering priority (i.e., left or right) when facing an obstacle as mentioned above, we customize the exploration policy by choosing different initial headings for each drone. This changes how the primary axis is initially selected.

%% file: sources/5-swarm.tex
\section{Swarm Coordination} \label{sec:swarm}

This section discusses the methodology used to enable swarm mapping and inter-drone radio communication.
In particular, it introduces the communication protocol and the types of messages the drones exchange with each other.
Furthermore, it explains how the drones coordinate within the swarm, the mapping mission, and how they exchange poses and scans to create a map.
Lastly, a scalability study is provided.

\subsection{Communication: Physical Layer}
The physical layer is provided by the Crazyflie Peer-to-Peer (P2P) interface enabled through its onboard radio transceiver. 
The P2P Application Programming Interface (API) provides basic packet-based communication primitives with best-effort delivery. 
No routing is implemented on this layer, meaning all packets are broadcast without any form of acknowledgment. 
Further, there is neither media access control nor flow control over the radio link. 
Since the layer provides completely unreliable package delivery, all link, network, and transport layer functionality must be implemented. 
Given the limited capabilities of nano-UAVs, it is desirable to leverage the existing hardware already available onboard rather than adding more complex communication transceivers that would result in more weight and additional power consumption.
Therefore, we rely on the available P2P communication and develop a robust and scalable communication protocol, which we present later in this section.
Since we also need to capture data from the swarm in a base-station computer (for monitoring and evaluation purposes only), we configure an additional \textit{bridge drone} that remains grounded, but relays all received packets to the computer over the built-in USB link. 

\subsection{Communication: Transport Layer}
To forgo the complexity and resource requirements of, for example, a TCP/IP stack, we designed a lightweight transport layer protocol to provide reliable message-based networking with broadcast and unicast capabilities.

In order to simplify the protocol, we make some assumptions about the communication and impose some limitations. 
Most importantly, each transmitter may only attempt to transmit a new packet when all intended recipients have acknowledged the previous packet. 
This significantly reduces the complexity of the receiver and provides some rudimentary flow control. 
Further, we assume that all drones are within communication range of each other. 
This allows us to forego the need to implement a mesh networking scheme that would be out of the scope of this work. Since synchronization between drones in our method is event-driven and no long-lasting data streams between drones are necessary, we recognize that message-based networking is more appropriate than connection-oriented networking. In particular, most messages are short, and large messages are transmitted infrequently and are not broadcast. Message-based communication in this form reduces complexity, as less state needs to be maintained. 

The basic principles of our proposed protocol are derived from ALOHAnet.
However, to reduce the memory footprint of each layer, the protocol merges the link, network, and transport layers, which allows packets to be evaluated without copying individual sections. Furthermore, this enables us to merge all headers for these layers into a 16-bit wide structure and increase the utilization of physical layer packets.

Since the physical layer packets carry at most 60 bytes of user data, messages must be split into individual packets and reassembled at the receiver. 
Given the first limitation we imposed above, reassembly is trivial. 
We use the 16-bit wide packet header of the following structure to implement the protocol: 4-bit \textit{source} and \textit{destination} field, \textit{acknowledge} bit, \textit{end} bit, \textit{tag} field, and \textit{sequence number} field. The source and destination refer to the addresses of the transmitter and receiver drone. A destination field of address \verb|0xF| indicates a broadcast, while the source field must not contain the broadcast address. The sequence number of each packet is an increasing unsigned integer that is used to deduplicate packets that have been retransmitted. When receiving a packet from another drone, the receiver responds with an empty packet with the acknowledge flag set. Once the sender receives the acknowledgment (ACK) packet from all targeted drones, it may send another message. Message boundaries are indicated by setting the end flag on the last packet of a message. The application layer uses the tag field to help identify the type of message received. No length field is included in the header since a lower layer header already provides this information.

\subsection{Communication: Message types}
In the application layer, we distinguish between four different types of messages that are each identified using the tag field in the protocol header.
Firstly, there is the \textit{pose update message} (PUM) -- 16 bytes: this message is used to synchronize the pose graph between drones. 
Each message contains a pose data structure and is broadcast to the entire swarm. 
The same message type is used both to register new poses and update existing poses. 
Since each node is identified by a globally unique identifier, these two scenarios can be distinguished by checking whether the pose identifiers already exist.
The next message type is the \textit{ToF scan request} (TSR) -- 4 bytes: this is a unicast message that is sent by the main drone to request the transfer of a ToF scan. 
The body of this message is a pose identifier associated with the scan that is being requested.
Paired with the ToF scan request, we introduce the \textit{ToF scan response} (SR) -- 1146 bytes: this message is sent in response to a ToF Scan Request message. The body contains the 2D points in the requested scan.
Lastly, we have the \textit{control messages} -- 16 bytes: this class of messages contains control flow instructions, such as the takeoff or landing commands sent by the base station through the bridge drone. 

\subsection{Mission Coordination}
Even if identical in terms of hardware, the drones can be classified by their functionality.
Firstly, we have the bridge drone that we introduced above.
Then, we have the flying drones in the swarm that explore the environment and acquire scans and poses.
Among the flying drones in the swarm, one drone is elected as the leading main drone.
Although all drones contribute to the mapping task, the main drone collects relevant scans, performs the ICP and SLAM computation to optimize the poses of the global graph, and propagates the results back to the swarm.
Despite applying a centralized approach to execute the SLAM algorithm, the architecture does not, in fact, require the main drone to store all of the data acquired by the other drones. 
Memory-intensive data, such as the individual ToF scans captured by other drones, can be loaded from the swarm on demand. 
This is important to ensure that this approach can scale with the size of the swarm, given the tight memory constraints of individual drones and bandwidth constraints of the entire swarm, and to increase the overall swarm robustness, avoiding single points of failure.
The pose graph is the only data shared and synchronized between all drones in the swarm formation, which is lightweight and scales linearly with the total length of all flight paths in the swarm.
Given that the pose graph is shared between all drones and all other data can be loaded on-demand, the architecture provides fault tolerance for the main drone. 
Should the main drone fail, any of the remaining drones are able to assume its role.

Before the mapping mission starts, all drones are placed in known positions.
When the bridge drone broadcasts the take-off message, all drones start exploring the environment according to the exploration policy introduced in Section~\ref{sec:autonomous-exploration}.
Whenever a drone reaches a way-point (i.e., every \SI{1}{\meter}), it adds a new pose in the graph and then broadcasts it to the swarm.
Furthermore, a new scan is acquired and stored in the external flash memory. 
For every new pose, the main drone checks whether it is within proximity of another pose, indicating that there should be sufficient overlap between the corresponding ToF scans to perform scan matching. Should this be the case, the main drone will request the relevant scan data from the appropriate drone, execute ICP, and add a virtual edge to the pose graph. 
At the end of the mission, the main drone can execute the SLAM algorithm and broadcast corrected pose information back to the swarm.
The pose graph is optimized at the end of the mission to simplify the evaluation, but due to the relatively low execution time of SLAM (i.e., several seconds) with respect to the mission time (i.e., several minutes), the optimization can also happen multiple times during the mission.
After SLAM is run, the optimized poses are used to correct the scans, and the map is obtained by merging all scans together.

\subsection{Scalability}

A performance and scalability analysis of the presented communication protocol is proposed. In particular, we consider the required bandwidth for broadcast communications with respect to the number of swarm agents, normalized with the inter-pose distance $d=$ \qty{1}{\metre} (also \qty{1}{\metre} in our work). 
Hence, we define the required swarm bandwidth $B_{s}(N,d)$ as a function of the number of agents $N$ and the pose-to-pose distance $d$ as shown in Equation~\ref{eq:scalability}. As introduced earlier in this section, \textit{PUM}, \textit{TSR}, and \textit{SR} are 16 bytes, 4 bytes, and 1146 bytes, respectively. Furthermore, \textit{ACK} is 2 bytes since an acknowledge requires sending a complete header without any payload.
$P_{sm}$ defines the probability two drones fly through the same location -- i.e., probability to require the scan-matching.
Normally, $\chi_{upd} \cdot (PUM_{tot}+f_{map}\cdot (ACK\cdot N))$ should also be added to $B_{s}(N,d)$ in Equation~\ref{eq:scalability} and represents the overhead for broadcasting the map from the main drone to the swarm.
However, $\chi_{upd}$ defines the map update rate, which is broadcast to all drones after running SLAM, and it is zero in our work since we only run SLAM once at the end of the mission.
$f_{map}$ and $f_{scan}$ are the packet fragmentation ratios -- i.e., the minimum integer number of 60 byte packets to send a map and a scan, respectively.
Since the size of a scan is 1146 bytes, $f_{scan} = 20$, while
$f_{map}$ depends on the environment, but it is bounded by the size of the flash.

\begin{equation}
\begin{split}
    B_{s}(N,d)=\underset{pose \ broadcasting}{\underbrace{\frac{N}{d} \left(  PUM + (ACK \cdot (N-1)) \right)}} + \\
    \underset{scan \ broadcasting}{\underbrace{N \cdot P_{sm} \cdot \left (  TSR + SR + f_{scan} \cdot ACK \right )}}~.
\end{split}
    \label{eq:scalability}
\end{equation}

Considering the practical experiments reported in Section~\ref{txt:results} with 2 or 4 nano-UAVs, the required bandwidth scanning a new pose every \qty{1}{\m} is \qty[per-mode=symbol]{4.05}{\kilo\bit\per\s} and \qty[per-mode=symbol]{8.24}{\kilo\bit\per\s}, respectively. Notably, the bandwidth requirement scales approximately linearly. Considering the Crazyflie P2P interface, the maximum measured bitrate in our swarm is \qty[per-mode=symbol]{64.1}{\kilo\bit\per\s}, which would be able to support a swarm composed of 20 agents. However, if a \qty[per-mode=symbol]{6}{\mega\bit\per\s} UWB communication would be used instead~\cite{polonelli2022open} -- commonly used in nano-UAV platforms -- the maximum swarm size could reach 500 with $d=\qty{1}{\m}$.

%% file: sources/6-results.tex
\section{Results}
\label{txt:results}

In this section, we provide a quantitative analysis of ICP and SLAM in terms of scan-matching accuracy, trajectory correction, and mapping accuracy.
All experiments are performed in our testing arena, equipped with a Vicon Vero 2.2 motion capture system (mocap) for ground-truth measurements of the drones' poses.
To assess the localization and mapping capabilities of our swarm, we build several mazes out of \qtyproduct{100 x 80}{\cm} chipboard panels. Figure~\ref{fig:mazes_multi} shows in-field acquired images with the mazes used in our evaluation.

\begin{figure*} [t]
\begin{centering}
\includegraphics[width=\linewidth]{figures/results/mazes_multi.pdf}
\par\end{centering}
\centering{}
\caption{Images taken in-field showing our evaluation environments. a) shows the experimental setup for the experiment presented in Section~\ref{sec:slam-results}. The mazes shown in b), c), and d) correspond to Experiments 1, 2, and 3 from Section~\ref{mapping-results}} \label{fig:mazes_multi}
\end{figure*}

\subsection{ICP Results}
In the following, we evaluate the scan matching accuracy in terms of rotation error and translation error.
In this scope, we place a drone in two nearby locations and acquire a scan in each.
Then, we execute onboard ICP and compute the relative transformation between the poses associated with the two.
Furthermore, with the aid of the mocap, we also compute the ground truth transformation since the mocap precisely measures the position and heading of the two poses.
To calculate the error, we use a method similar to the one proposed by Pomerleau et al.~\cite{pomerleau2013comparing}. 
We express the ground truth transformation and the one provided by ICP in homogenous coordinates, which we note $\mat{T}_{GT}$ and $\mat{T}_{ICP}$.
Since homogenous coordination allows unifying rotation and translation in the same transformation, we calculate the ICP error using Equation~\ref{eq:icp-error}.
Specifically, the translation error is calculated as $e_t=\lVert \mathit{\Delta}\vec{t} \rVert$, where $\mathit{\Delta}\vec{t}$ comprises the translation error on both $X$ and $Y$ axes.
The rotation error is retrieved from the rotation matrix as $e_R=\cos^{-1}( \mathit{\Delta}\mat{R}_{00})$.
\begin{equation}
    \mathit{\Delta}\mat{T}
    =
    \begin{bmatrix}
    \mathit{\Delta}\mat{R} & \mathit{\Delta}\vec{t} \\
    \vec{0}^\top & 1 \\
    \end{bmatrix}
    =
    \mat{T}_{ICP} \mat{T}_{GT}^{-1}
    \label{eq:icp-error}
\end{equation}
\begin{figure}[t]
    \centering
    \begin{subfigure}{0.49\columnwidth}
        \centering
        \includegraphics[width=\linewidth]{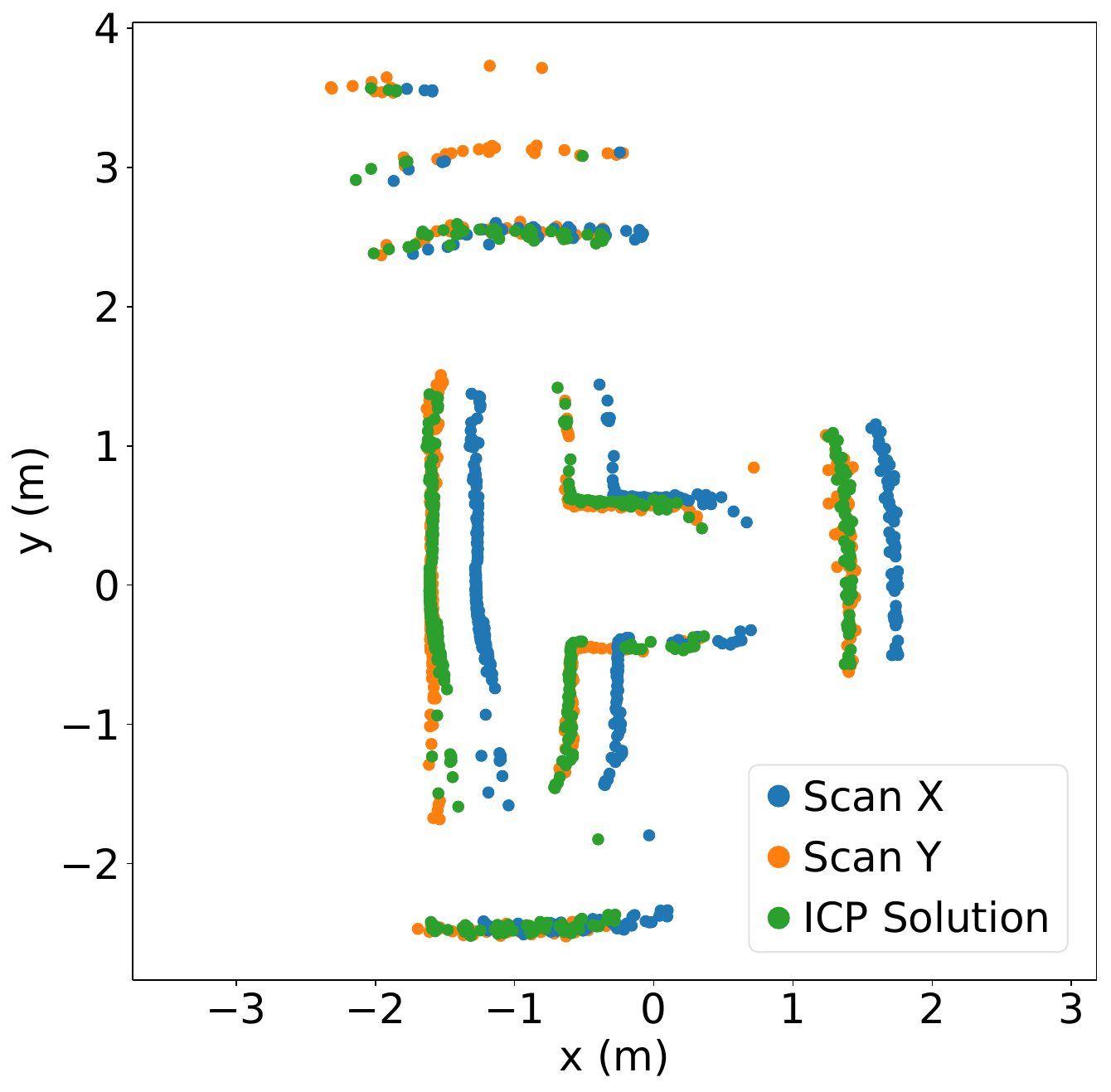}
        \caption{$e_t$ = \qty{0.7}{\cm}, $e_R$ = \ang{0.98}
        \label{fig:icp-a}}
    \end{subfigure}
    \begin{subfigure}{0.49\columnwidth}
        \centering
        \includegraphics[width=\linewidth]{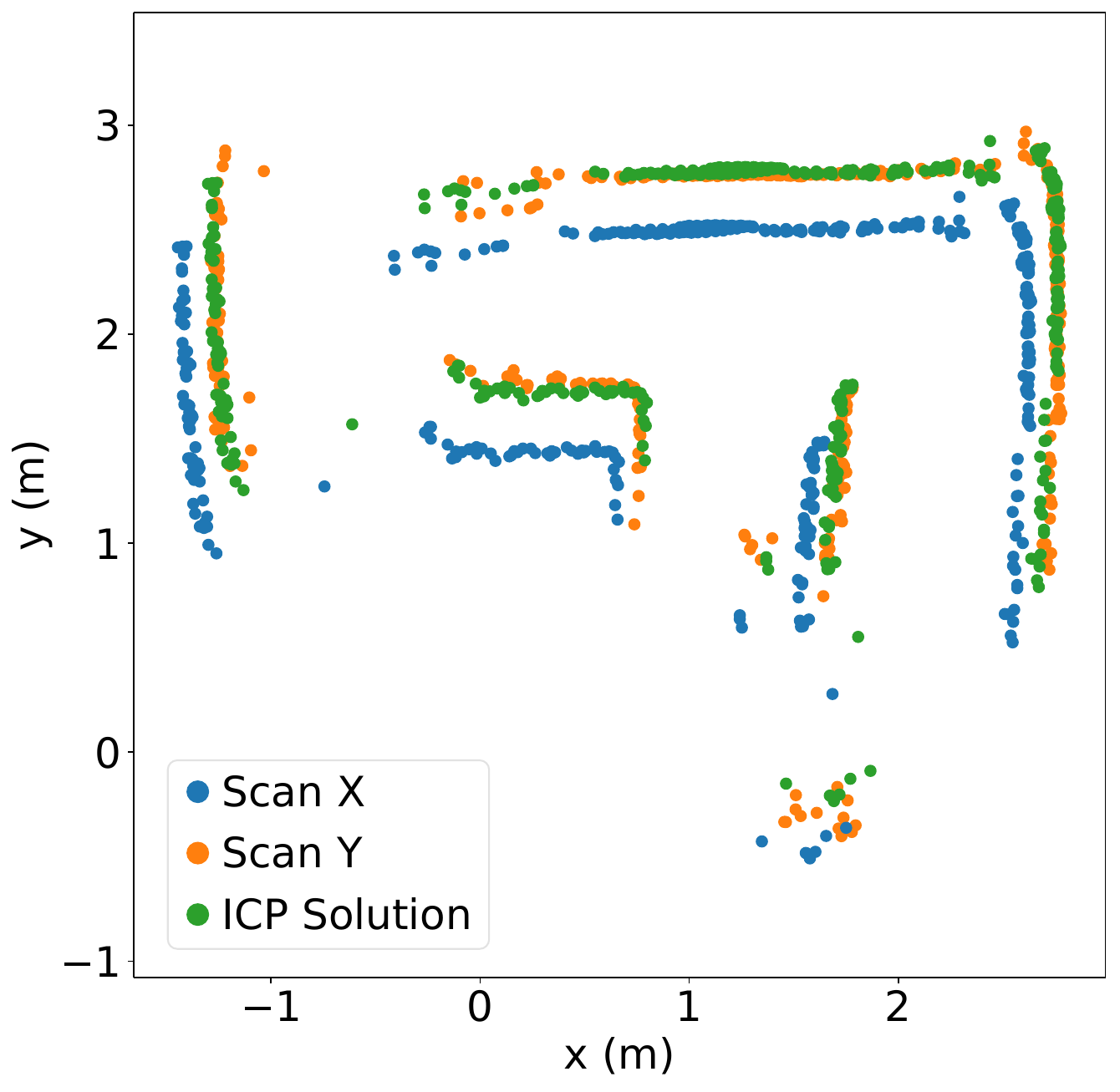}
        \caption{$e_t$ = \qty{3.2}{\cm}, $e_R$ = \ang{0.56}
        \label{fig:icp-b}}
    \end{subfigure}
    \caption{Translation and rotation error for the onboard computed ICP solutions using empirical ToF scans.}
    \label{fig:icp-eval}
\end{figure}

\begin{table} [b]
\caption{Accuracy of ICP as a function of the number of scan frames.}
\label{tab:icp-vs-frames}
\centering
\begin{tabular}{lllllll} 
\hline\hline
 & Frames      & 3    & 6    & 9    & 12   & 15    \\ 
\hline
\multirow{2}{*}{Figure~\ref{fig:icp-a}} & $e_t$ (\qty{}{\centi\meter}) & 33.8  & 8.6  & 1.5  & 0.6  & 0.7  \\
                         & $e_R$ (deg)   & 2.21 & 1.34 & 1.39 & 1.02 & 0.98  \\ 
\hline
\multirow{2}{*}{Figure~\ref{fig:icp-b}} & $e_t$ (\qty{}{\centi\meter}) & 6.3  & 4.4  & 3.6  & 3.2  & 3.2   \\
                         & $e_R$ (deg)   & 1.44 & 1.54 & 1.30 & 0.57 & 0.56  \\ 
\hline
                         & Memory (\qty{}{\byte})        & 768 & 1536 & 2304 & 3072 & 3840   \\
\hline\hline
\end{tabular}
\end{table}

Figure~\ref{fig:icp-eval} presents the in-field results for two scenarios, where the scans are acquired in a corridor intersection (i.e., Figure~\ref{fig:icp-a}) and in a corner (i.e., Figure~\ref{fig:icp-b}).
In both cases, the 2D points of each scan are represented in blue and orange, respectively.
Green represents the scan created by applying the ICP transformation to the scan $X$ -- which in both cases approximately matches scan $Y$.
In the first experiment depicted in Figure~\ref{fig:icp-a}, the ICP algorithm achieves a highly accurate solution with a translation error of only \SI{0.7}{\centi\meter} and a rotation error of \ang{0.98}.
In contrast, the translation error in the second experiment from Figure~\ref{fig:icp-b} is slightly higher (i.e., \SI{3.2}{\centi\meter}) while the rotation error is \ang{0.56}.
The increased translation error is most likely due to the poorer texture, such as the smaller number of corners in the scan.
Furthermore, on the bottom part of Figure~\ref{fig:icp-b}, one could notice some artifacts.
Since the scan generation is based on the drone's state estimate, errors in this estimate can lead to outliers in the scan.
Acquiring a scan requires the drone to spin around its z-axis, and the state estimate that mainly relies on the onboard optical flow sensor is not very stable during spinning.

As firstly mentioned in Section~\ref{sec:algorithms}, we use a scan size of 480 2D points.
In the following, we justify this choice by analyzing how the rotation and translation error achieved by ICP changes with the scan size.
Since the scan is obtained by stacking multiple scan frames, we perform this analysis as a function of the number of scan frames.
Therefore, we vary the frame count in the range of 3 -- 15 with a step of 3 and evaluate the translation and rotation errors.
Furthermore, this evaluation is performed for both experiments depicted in Figure~\ref{fig:icp-a} and Figure~\ref{fig:icp-b}, respectively.
The results are shown in Table~\ref{tab:icp-vs-frames} and prove that using more than 12 scan frames does not bring any improvement in terms of accuracy for both rotation and translation.
Beyond this value, the change in rotation and translation is below 5\%.
However, we chose the value of 15 scan frames to have more robustness and account for possible corner cases.
Table~\ref{tab:icp-vs-frames} also provides the maximum size of one individual scan, which scales linearly with the size of the scans.
For a scan consisting of 15 scan frames, the scan size is \qty{3840}{\byte} as every scan frame contains 32 2D points (\qty{8}{\byte} per 2D point).

\begin{figure*}[t]
    \centering
        \begin{subfigure}[b]{0.3\linewidth}
        \centering
        \includegraphics[width=\linewidth]{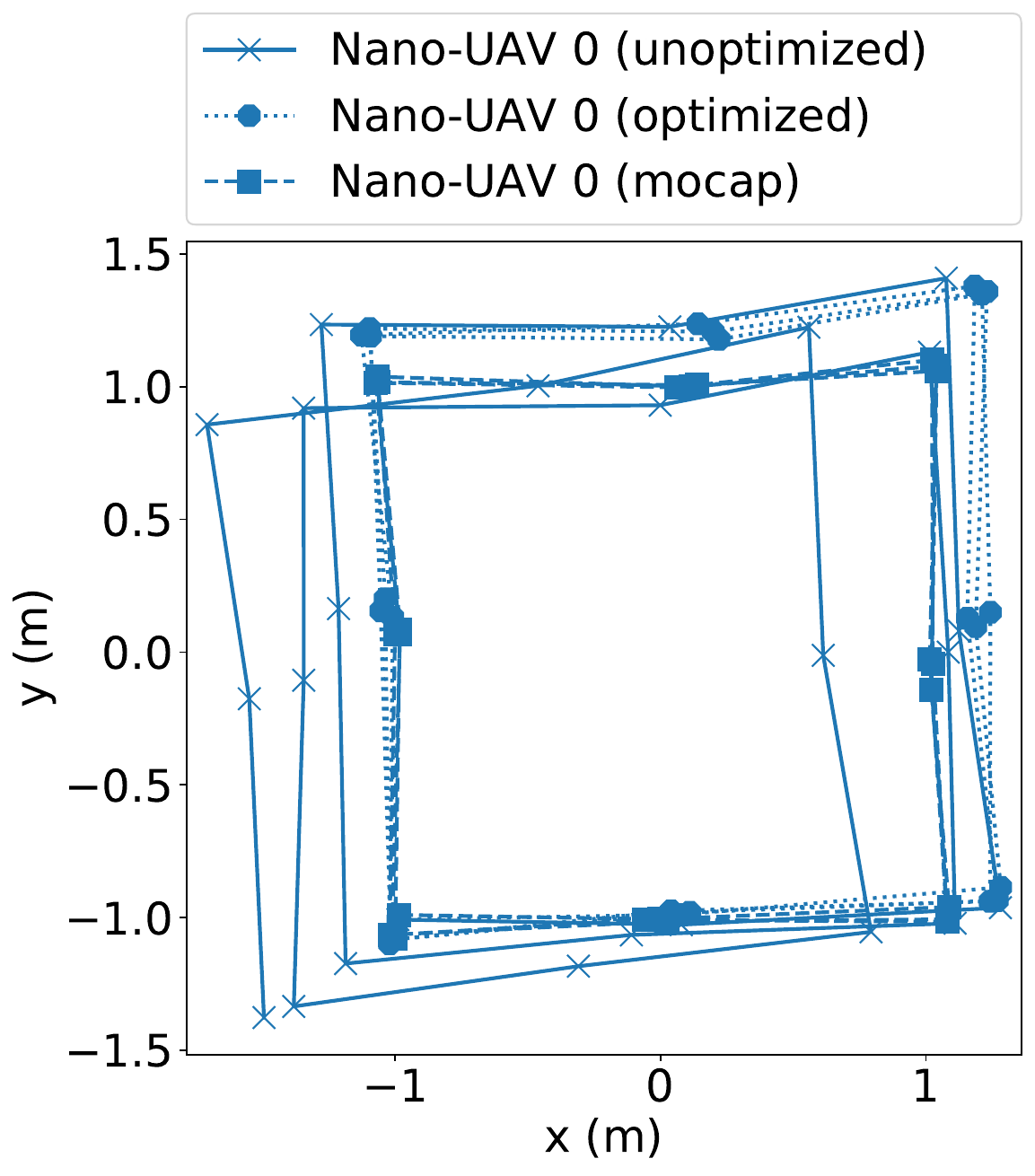}
        \caption{Trajectories and poses.
        \label{fig:slam-poses}}
    \end{subfigure}\hfill%
    \begin{subfigure}[b]{0.3\linewidth}
        \centering
        \includegraphics[width=\linewidth]{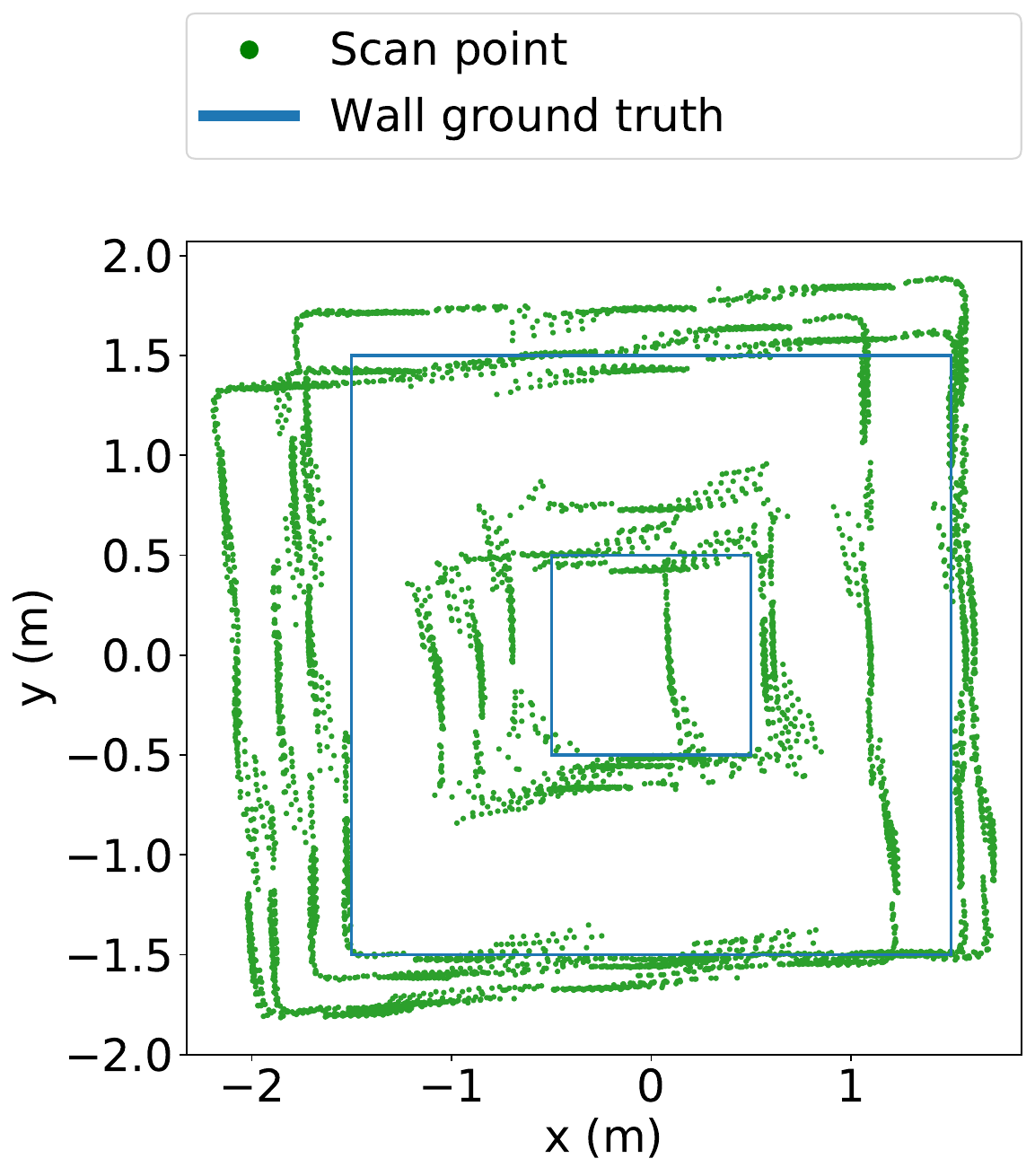}
        \caption{Map before optimization.
        \label{fig:slam-a}}
    \end{subfigure}\hfill%
    \begin{subfigure}[b]{0.3\linewidth}
        \centering
        \includegraphics[width=\linewidth]{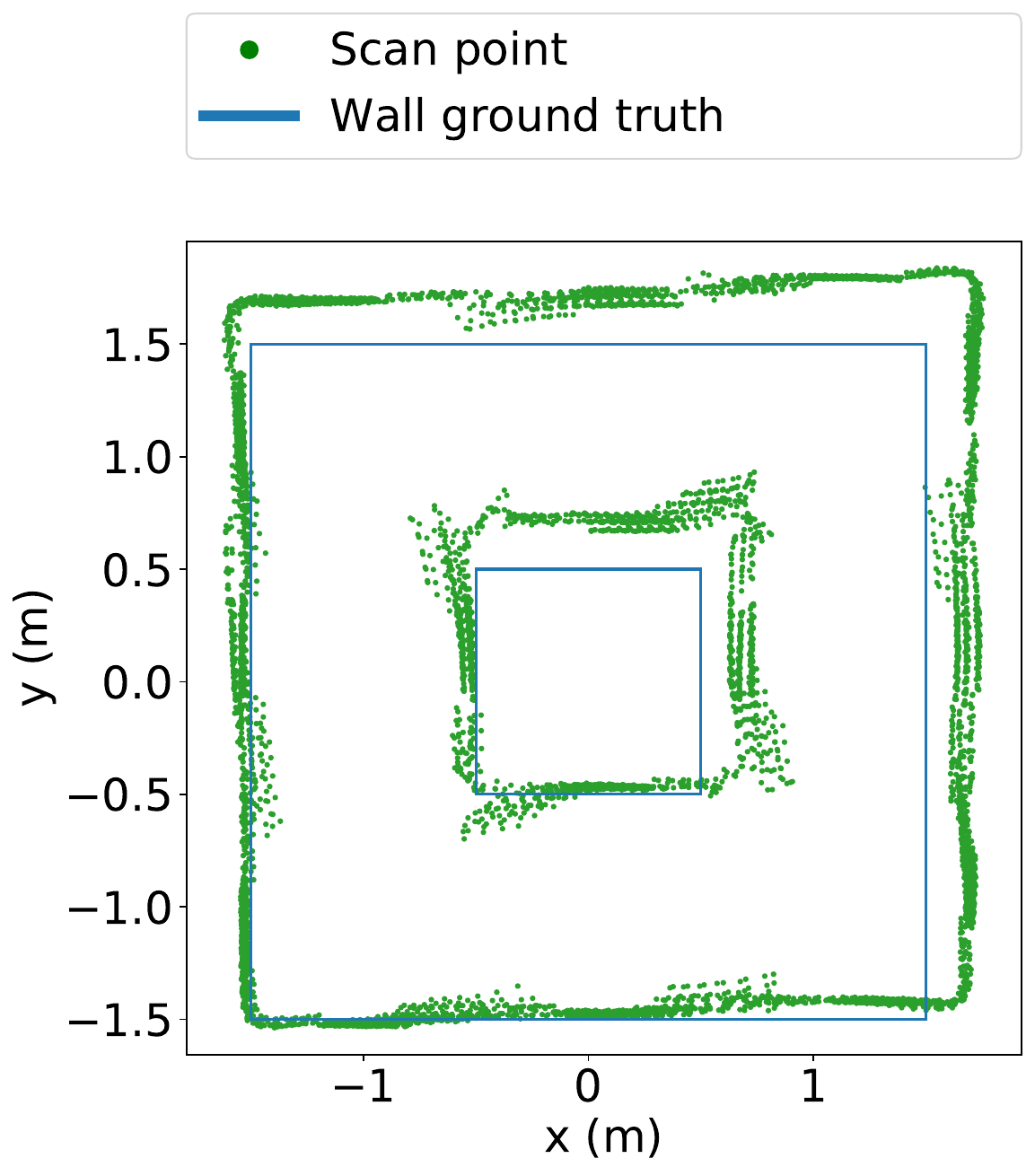}
        \caption{Map after optimization.
        \label{fig:slam-b}}
    \end{subfigure}
    \caption{Mapping a square maze with a single nano-UAV.
    \label{fig:slam-exp}}
\end{figure*}

\subsection{SLAM Results} \label{sec:slam-results}
We recall that generating accurate maps requires accurate trajectory estimation, as the drone's trajectory represents the foundation for projecting the distance measurements in the world frame, as shown in Section~\ref{sec:algo-scans}.
Therefore, we proceed by assessing the performance of the SLAM in correcting the trajectory errors and then generate the map and evaluate its accuracy.
The experiments in this section only involve one drone agent.
To evaluate the performance of the SLAM algorithm in correcting the trajectory, we check how close the estimated poses are to the ground truth poses provided by the mocap. Given a set of estimated poses $\vec{x}_1 \ldots \vec{x}_n$, the corresponding set of ground truth poses is $\vec{x}_1^{GT} \ldots \vec{x}_n^{GT}$.
We mention that we use a reduced representation of the poses, each consisting only of the x and y coordinates.
The pose estimation Root-Mean-Squared-Error (RMSE) of the optimized poses with respect to the ground truth is computed as in Equation~\ref{eq:rmse_poses}.
\begin{equation} \label{eq:rmse_poses}
    RMSE_{poses} = \sqrt{\frac{\sum_{i=1}^{n}{\lVert \vec{x}_i - \vec{x}_i^{GT} \rVert^2}}{n}}
\end{equation}

To evaluate the pose correction performance, we proceed with an experiment where the drone is autonomously flying in a simple maze of square shape, with another square obstacle in the middle, just as shown in Figure~\ref{fig:slam-a} or \ref{fig:slam-b} in blue or in Figure~\ref{fig:mazes_multi}-a.
The drone starts flying from the bottom left corner of the square and completes the loop three times.
It flies at a constant height of \SI{0.6}{\meter} and uses an exploration speed of $v_{exp}=\SI{0.8}{\meter / \second}$, acquiring scans in each waypoint.
Whenever it revisits a waypoint, it uses ICP scan matching to create SLAM constraints with respect to the scans acquired in the first trajectory loop.
After flying the three loops, the poses are optimized by running SLAM.

Figure~\ref{fig:slam-poses} shows the three trajectories: before SLAM (i.e., unoptimized), after SLAM (i.e., optimized), and the ground truth.
The markers indicate the location of the poses while the lines interpolate between poses.
The solid line indicates the unoptimized trajectory -- i.e., as determined by the internal state estimator, while the dashed line represents the ground truth measured by the mocap. 
Furthermore, the dotted line shows the optimized flight path, which is closer to the ground truth.
Specifically, the RMSE of the unoptimized and optimized poses is \SI{34.6}{\centi\meter} and \SI{19.8}{\centi\meter}, respectively.

We further evaluate the accuracy of the map, which is generated by putting together all the scans.
Firstly, we propose a metric that quantifies the mapping accuracy.
We define walls as lines that span between two endpoints. 
If the point is above the wall, we define the distance of the point to the wall as the shortest distance to the line intersecting the endpoints. Otherwise, we define it as the distance to the closest endpoint. 
Given a set of walls $W$ and a set of 2D points in the map $\vec{p}_1 \ldots \vec{p}_n$, we define the mapping RMSE as in Equation~\ref{eq:rmse_map}.
We apply this metric to the map generated with the unoptimized poses and show the results in Figure~\ref{fig:slam-a}, which leads to a mapping RMSE of \SI{25.1}{\centi\meter}.
Re-projecting the scans using the optimized poses results in a corrected map shown in Figure~\ref{fig:slam-b}.
The mapping RMSE of the corrected map is \qty{16}{\cm}, proving that applying the correction based on ICP and SLAM improves the mapping accuracy by about 35\%.
Also, this time, we observe some artifacts, particularly at the corners of the walls, which are caused by inaccurate state estimation during the yaw rotation of a scan.
Note, however, that the map appears scaled by some constant factor. 
This is due to the drone's state estimator, which consistently overestimates distances. 
This problem can be mitigated by performing odometry calibration, but it is out of scope and left for future work.
\begin{equation} \label{eq:rmse_map}
    RMSE_{map} = \sqrt{\frac{\sum_{i=1}^{n}{(\min_{\forall w \in W} dist(w, \vec{p}_i))^2}}{n}}
\end{equation}

The experiment was run with an exploration velocity $v_{exp}=\SI{0.8}{\meter / \second}$.
If we run the same experiment again but with $v_{exp}=\SI{0.2}{\meter / \second}$, we obtain a pose estimation RMSE of \SI{20.4}{\centi\meter} and \SI{14.5}{\centi\meter} for the unoptimised and optimized poses, respectively.
Note that even without any SLAM optimization, the pose estimation RMSE is smaller at lower velocity, indicating that the odometry drift is strongly related to the drone's velocity. 

\begin{figure*}[t]
    \begin{subfigure}{0.23\linewidth}
        \centering
        \includegraphics[width=\linewidth]{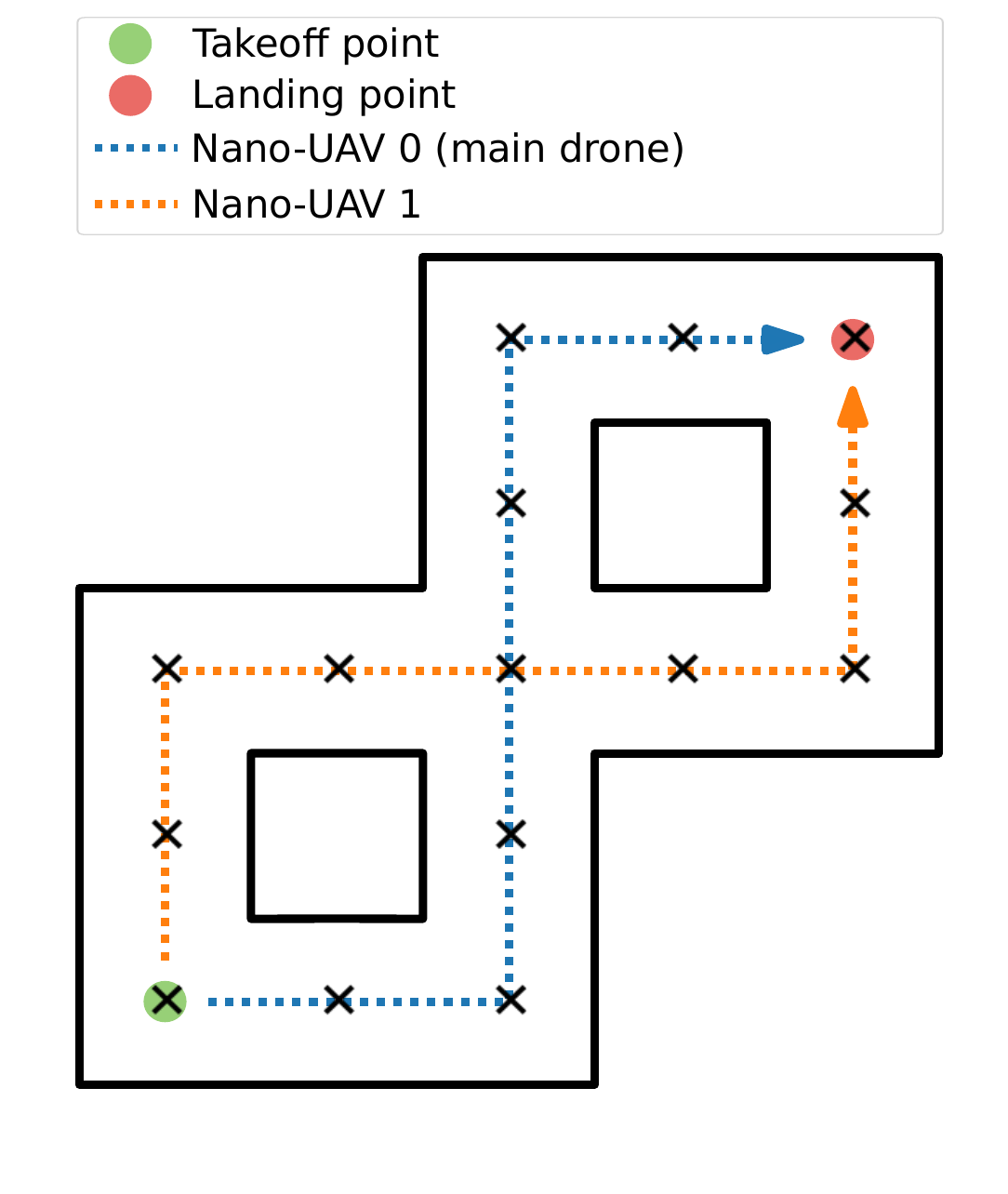}
        \caption{Maze layout.
        \label{fig:mapping-exp1}}
    \end{subfigure}\hfill%
    \begin{subfigure}{0.23\linewidth}
        \centering
        \includegraphics[width=\linewidth]{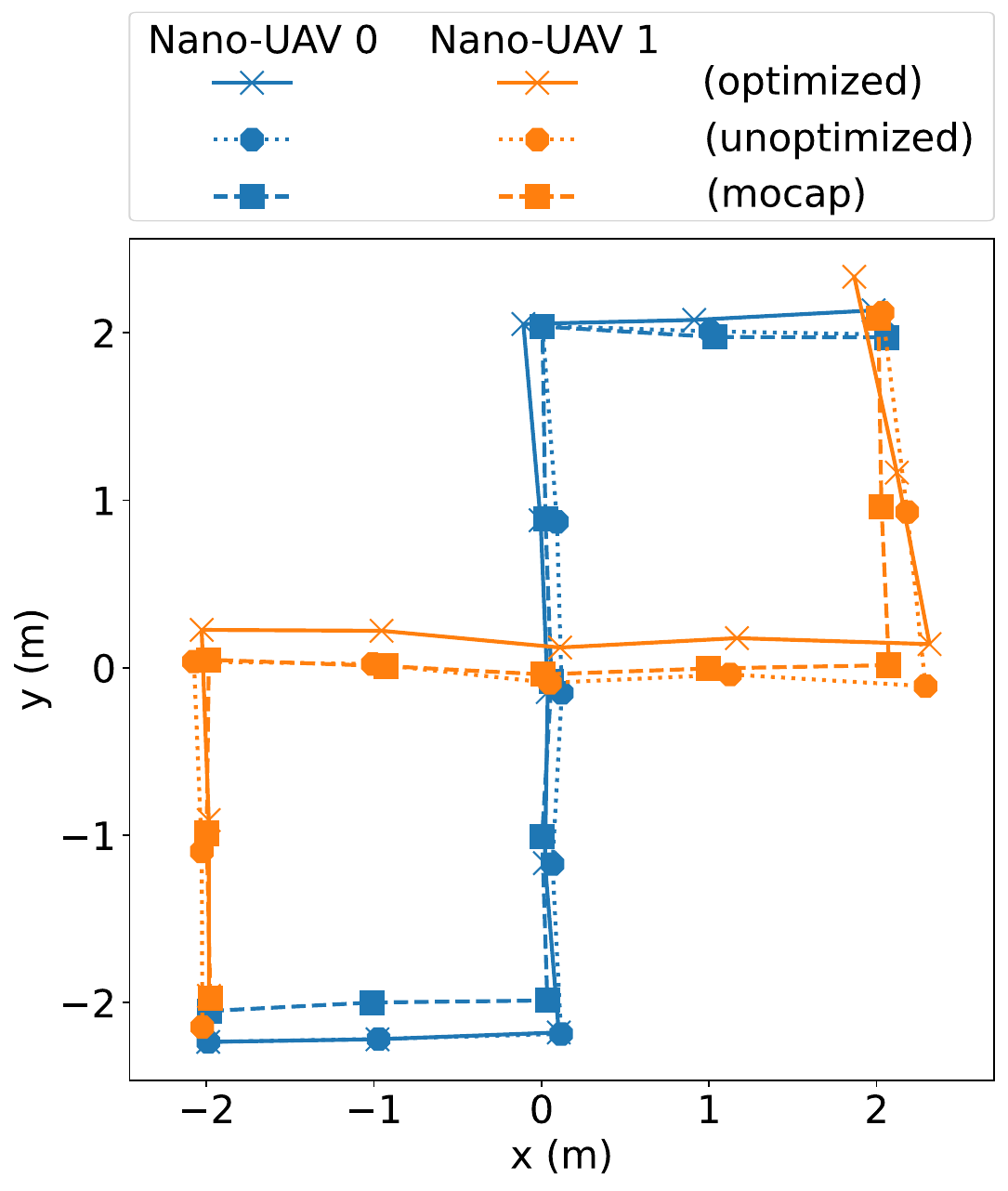}
        \caption{The trajectories.
        \label{fig:exp1-poses}}
    \end{subfigure}\hfill%
    \begin{subfigure}{0.23\linewidth}
        \centering
        \includegraphics[width=\linewidth]{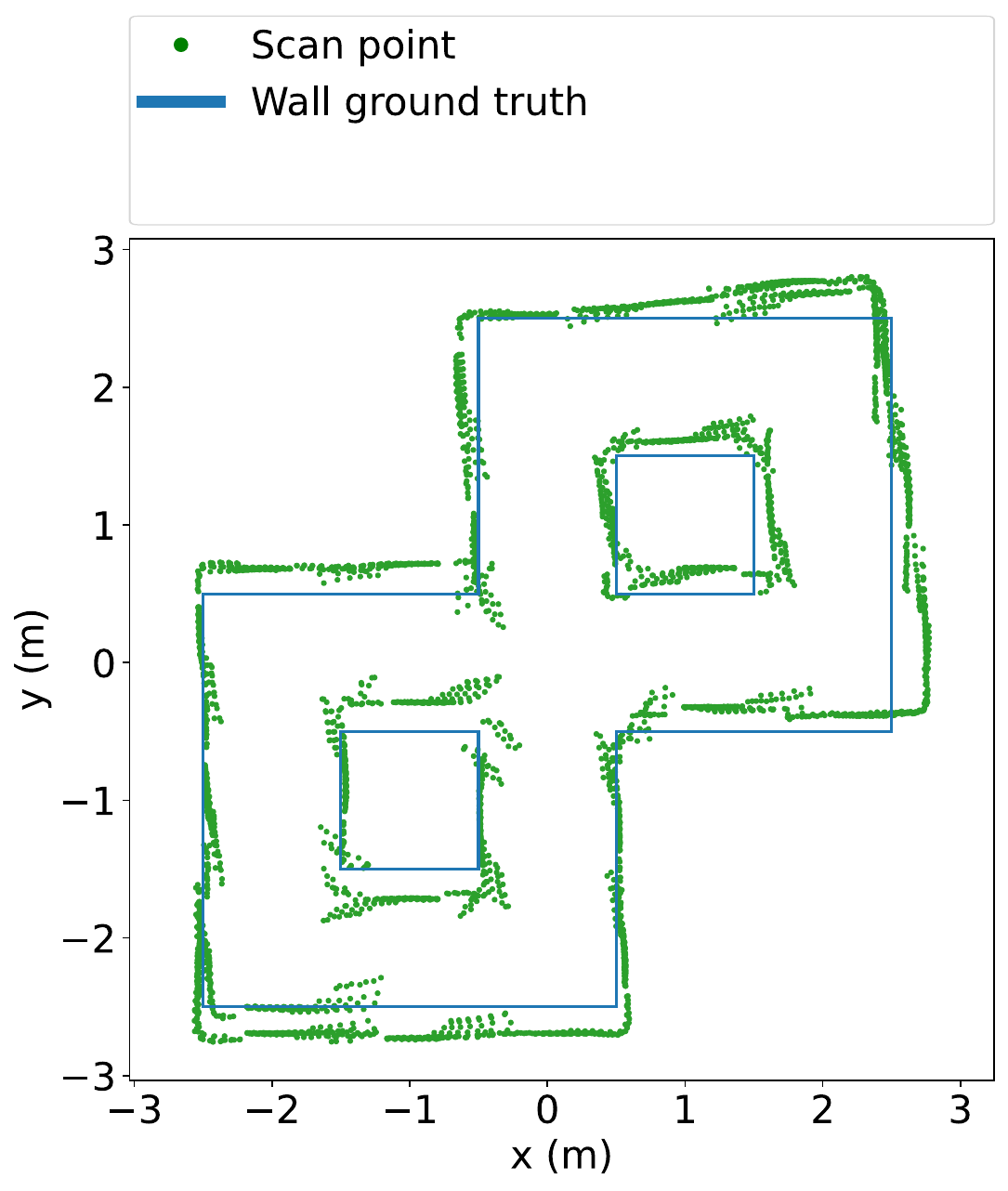}
        \caption{Unoptimized map.
        \label{fig:exp1-map-before}}
    \end{subfigure}\hfill%
    \begin{subfigure}{0.234\linewidth}
        \centering
        \includegraphics[width=\linewidth]{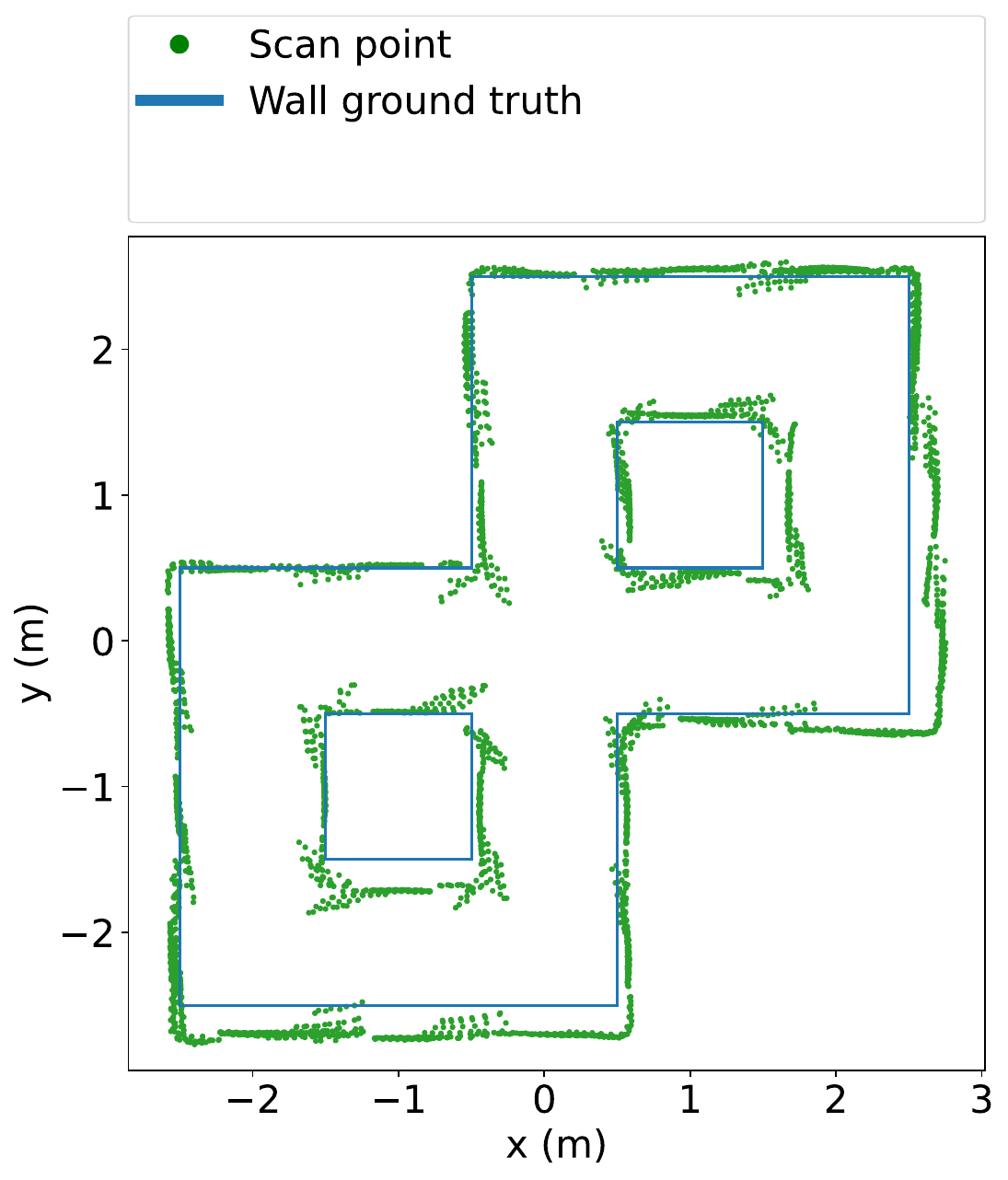}
        \caption{Optimized map.
        \label{fig:exp1-map-after}}
    \end{subfigure}
    \caption{First experiment: mapping experiment with a swarm of two nano-UAVs.
    \label{fig:exp1}}
\end{figure*}

\subsection{Mapping Results} \label{mapping-results}
So far, in this section, we evaluated the performance of ICP and SLAM, and we proved the benefit of applying SLAM for correcting the trajectory and generating a map with one drone.
In the following, we focus on distributed mapping and evaluate the mapping capabilities when a swarm of drones is employed.
To prove the generalization capabilities of our system, we performed three experiments, each consisting of mapping a different maze.
We recall that the swarm-based experiments imply having one main drone, which collects the poses and scans from the other drones, runs the optimization, and then sends out the corrected poses.
We proved in Section~\ref{sec:slam-results} that the exploration velocity impacts the odometry accuracy.
However, since a very low velocity of the drones would result in a large mapping time, we set $v_{exp}=\SI{0.4}{\meter / \second}$ for the main drone and a larger velocity $v_{exp}=\SI{0.8}{\meter / \second}$ for the other drones as a trade-off between mapping accuracy and time.

\begin{table}[b]
    \centering
        \caption{Pose estimation and mapping accuracy of the second and third experiment.
    \label{tab:exp2-rmse}
    \label{tab:exp3-rmse}}
    \begin{tabular}{ ccccc } 
        \hline
        \hline
        & \multicolumn{4}{c}{Second experiment}  \\
        \cline{2-5}
        \multirow{2}{*}{Drones} & \multicolumn{2}{c}{Pose estimation RMSE} & \multicolumn{2}{c}{Mapping RMSE} \\
        \cline{2-5}
        & no-SLAM & SLAM & no-SLAM & SLAM \\
        \hline
        2 & \qty{45.5}{\cm} & \qty{20.6}{\cm} & \qty{30.9}{\cm} & \qty{13.1}{\cm} \\ 
        4 & \qty{28.6}{\cm} & \qty{16.5}{\cm} & \qty{22.5}{\cm} & \qty{14.5}{\cm} \\ 
        \hline
        & \multicolumn{4}{c}{Third experiment}  \\
        \hline
        2 & \qty{34.2}{\cm} & \qty{20.1}{\cm} & \qty{25.3}{\cm} & \qty{15.9}{\cm} \\ 
        4 & \qty{21.8}{\cm} & \qty{15.1}{\cm} & \qty{17.9}{\cm} & \qty{12.7}{\cm} \\
        \hline
        \hline
    \end{tabular}
\end{table}

\subsubsection{First Experiment}
The scenario of the first experiment is shown in Figure~\ref{fig:mazes_multi}-b and described in Figure~\ref{fig:mapping-exp1}.
The green and red markings denote the takeoff and landing locations for all drones, which are connected by the expected flight paths of each drone (dashed lines). 
The crosses mark the expected locations of the poses acquired during autonomous exploration. 
The unoptimized and optimized trajectories are shown in Figure~\ref{fig:exp1-poses} along with the ground truth poses for both drones -- shown in blue and orange.
Indeed, the pose estimation RMSE for the main drone is \SI{15.9}{\centi\meter}, and almost \qty{31}{\percent} higher for the other drone (i.e., \SI{20.8}{\centi\meter}).
The SLAM optimization can improve the situation significantly, reducing the overall pose estimation RMSE for both drones by about \qty{25}{\percent}.
Inspecting Figure~\ref{fig:exp1-poses} once again, we can see that the optimized poses (orange dotted line) for drone 1 are now close to the ground truth (orange dashed line). 
Similarly, comparing the generated map from Figure~\ref{fig:exp1-map-before} to Figure~\ref{fig:exp1-map-after} shows a clear improvement, with the mapping RMSE of the point cloud improving by \qty{19}{\percent} from \SI{14.4}{\centi\meter} to \SI{11.6}{\centi\meter}. 
The blue lines indicate the ground truth of where the walls are located.
    
\subsubsection{Second Experiment} \label{sec:second-exp}
\begin{figure*}[t]
    \centering
    \begin{subfigure}[b]{0.23\linewidth}
        \centering
        \includegraphics[width=\linewidth]{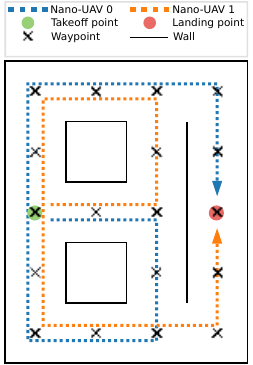}
        \caption{2 UAVs (Exp. 2)
        \label{fig:exp2-two-drones-maze}}
    \end{subfigure}%
    \begin{subfigure}[b]{0.23\linewidth}
        \centering
        \includegraphics[width=\linewidth]{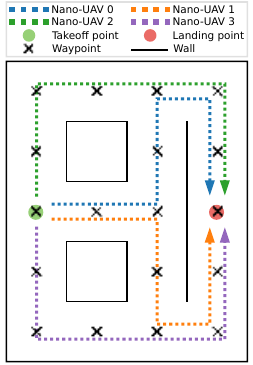}
        \caption{4 UAVs (Exp. 2)
        \label{fig:exp2-four-drones-maze}}
    \end{subfigure}
    \begin{subfigure}[b]{0.23\linewidth}
        \centering
        \includegraphics[width=\linewidth]{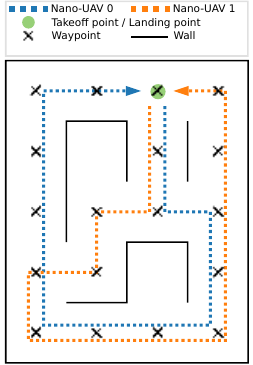}
        \caption{2 UAVs (Exp. 3)
        \label{fig:exp3-two-drones-maze}}
    \end{subfigure}
    \begin{subfigure}[b]{0.23\linewidth}
        \centering
        \includegraphics[width=\linewidth]{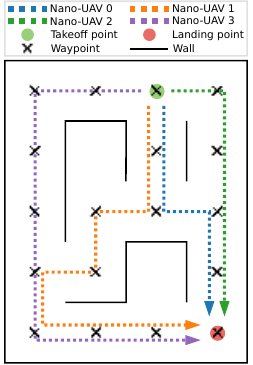}
        \caption{4 UAVs (Exp. 3)
        \label{fig:exp3-four-drones-maze}}
    \end{subfigure}
    \caption{The maze layouts for the second (a and b) and third experiment (c and d)}.
    \label{fig:exp1-maze}
\end{figure*}

\begin{figure*}[t]
    \centering
    \begin{subfigure}[b]{0.23\linewidth}
        \centering
        \includegraphics[width=\linewidth]{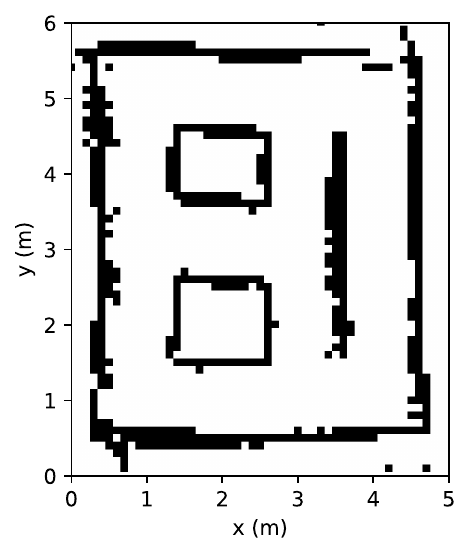}
        \caption{2 UAVs (Exp. 2)
        \label{fig:exp2-two-drones-binary-map}}
    \end{subfigure}
    \begin{subfigure}[b]{0.23\linewidth}
        \centering
        \includegraphics[width=\linewidth]{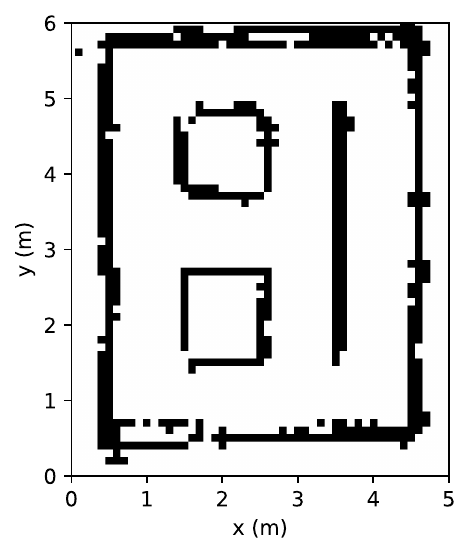}
        \caption{4 UAVs (Exp. 2)
        \label{fig:exp2-four-drones-binary-map}}
    \end{subfigure}
    \begin{subfigure}[b]{0.23\linewidth}
        \centering
        \includegraphics[width=\linewidth]{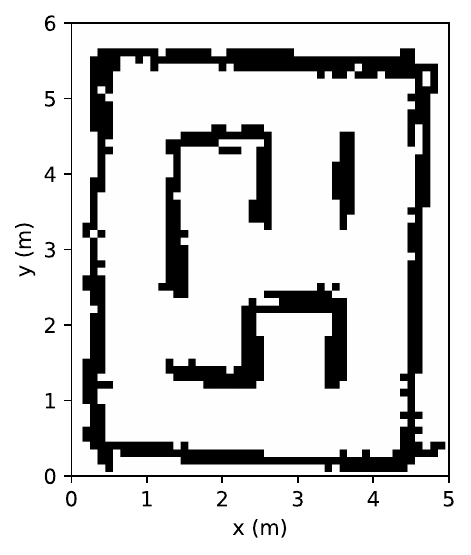}
        \caption{2 UAVs (Exp. 3)
        \label{fig:exp3-two-drones-binary-map}}
    \end{subfigure} 
    \begin{subfigure}[b]{0.23\linewidth}
        \centering
        \includegraphics[width=\linewidth]{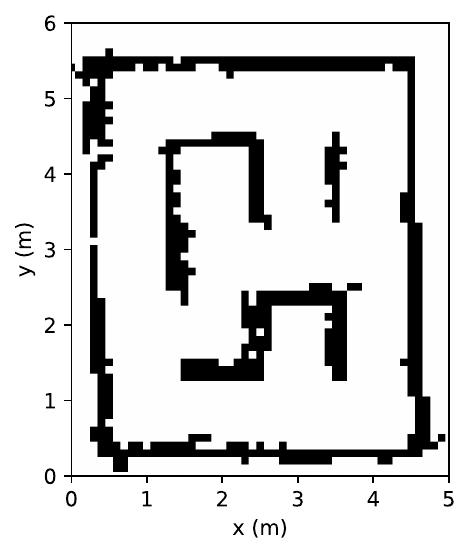}
        \caption{4 UAVs (Exp. 3)
        \label{fig:exp3-four-drones-binary-map}}
    \end{subfigure}
    \caption{The binary occupancy grid maps for the second (a and b) and third experiment (c and d).
    \label{fig:exp1-binary-map}}
\end{figure*}

In the second experiment (Figure~\ref{fig:mazes_multi}-c), we demonstrate the system's scalability as a swarm, evaluating not only the trajectory and mapping accuracy but also how the mapping time changes with the number of drones. 
For this scope, we map the same layout twice using swarms of two and four drones. 
The layout, as well as the intended flight paths, are visualized in Figure~\ref{fig:exp2-two-drones-maze} for two drones and in 
Figure~\ref{fig:exp2-four-drones-maze} for the scenario with four drones.
The main drone is always \textit{drone 0}.
Table~\ref{tab:exp2-rmse} lists the accuracy of the pose estimations and mapping with and without SLAM optimization for each swarm configuration. 
In general, the accuracy of pose estimations and mapping improves with more drones due to the shorter flight path per drone required to map the same area.
On average, the SLAM optimization approximately halves the pose estimation error.
The reduction in the mapping RMSE when SLAM is employed is 58\% and 36\% for the situation with two and four drones, respectively.
While the RMSE of the corrected map is about the same for two and four drones, the mapping time is heavily reduced, from \qty{5}{\minute} \qty{28}{\second} with two drones to \qty{2}{\minute} \qty{38}{\second} with four drones.
While keeping the same takeoff and landing location, we can cover the same area in less than half the time with four drones compared to two drones.
This result motivates the need for a drone swarm and proves the scalability potential.
In the first experiment, we provided the generated map in a dense representation (i.e., Figure~\ref{fig:exp1}).
Now we use the approach presented in~\cite{thrun2002probabilistic} to convert the dense map to an occupancy grid map with a resolution of \SI{10}{\centi\meter}, which we directly show in Figure~\ref{fig:exp2-two-drones-binary-map} and Figure~\ref{fig:exp2-four-drones-binary-map}.
We chose this representation as it is more meaningful for enabling future functionalities such as path planning.

\subsubsection{Third Experiment}
In this experiment, the objective is the same as in the second one, but the maze layout differs, as shown in Figure~\ref{fig:mazes_multi}-d.
The layout and the drones' trajectories are presented in Figure~\ref{fig:exp3-two-drones-maze} and Figure~\ref{fig:exp3-four-drones-maze} for two drones and four drones, respectively.
For this experiment, the maze layout is slightly more complicated to better approximate the diverse shapes of indoor environments. 
In this layout, it is not possible to use the same takeoff and landing locations for both swarm configurations. 
Instead, the drones in the two-drone swarm return to their starting point, while the drones in the four-drone swarm finish their flight path in a different location.
Table~\ref{tab:exp3-rmse} presents the pose estimation and mapping RMSE.
We observe an improvement in the pose estimation of 41\% for the two-drone case and 31\% for the four-drone case when using the SLAM optimization, while the mapping RMSE is reduced by 37\% and 29\%, respectively.
Due to the particularly long flight paths of drones 1 and 3 in the four-drone swarm, this configuration only benefits from a \qty{20}{\percent} mission time advantage in this layout, as the mission times are \qty{3}{\minute} \qty{7}{\second} and \qty{2}{\minute} \qty{30}{\second} for the cases with two drones and four drones, respectively.
We further remark that the accuracy figures when using SLAM optimization are similar to those observed in the previous experiment, as seen in Table~\ref{tab:exp2-rmse}.
As for the previous experiment, we provide the resulting occupancy grid maps, shown in Figure~\ref{fig:exp3-two-drones-binary-map} and Figure~\ref{fig:exp3-four-drones-binary-map}.
%


\subsection{Performance}
\begin{figure}[b]
    \centering
    \begin{subfigure}{0.49\columnwidth}
        \centering
        \includegraphics[width=\linewidth]{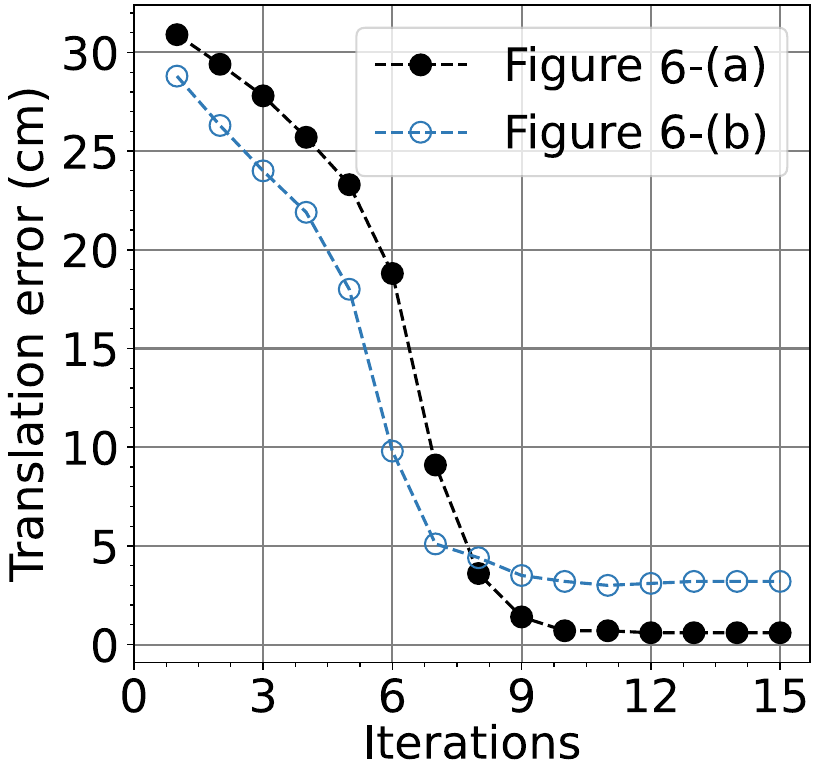}
        \caption{Translation error
        \label{fig:icp-iter-a}}
    \end{subfigure}
    \begin{subfigure}{0.49\columnwidth}
        \centering
        \includegraphics[width=\linewidth]{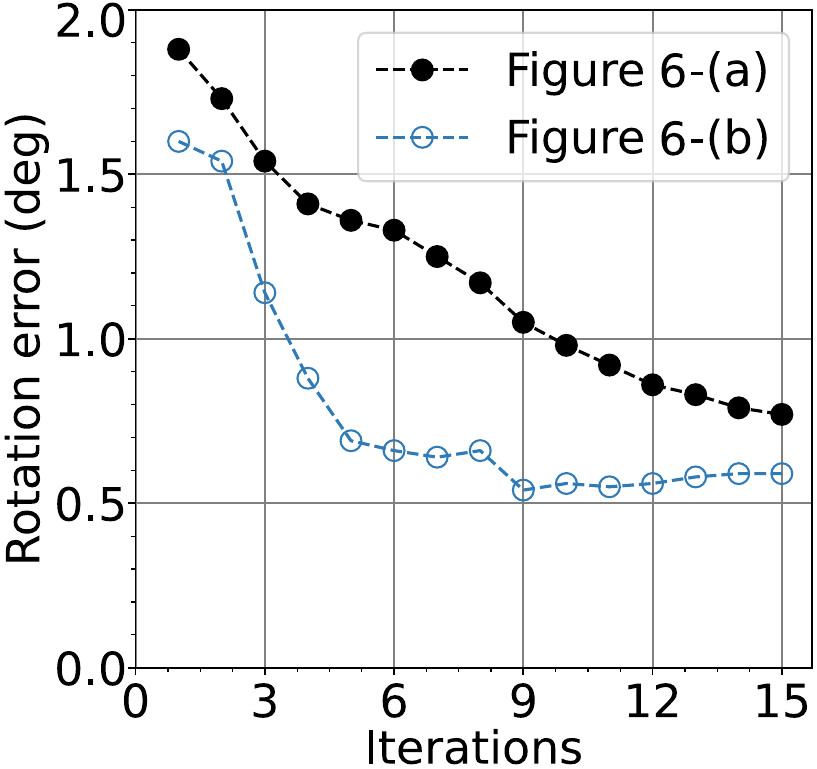}
        \caption{Rotation error
        \label{fig:icp-iter-b}}
    \end{subfigure}
    \caption{The ICP error as a function of the number of iterations.}
    \label{fig:icp-iter}
\end{figure}

As its name suggests, ICP is an iterative algorithm whose execution time scales linearly with the chosen number of iterations and quadratically with the number of 2D points in a scan -- due to the correspondence calculation as explained in Section~\ref{sec:algorithms}.
Since we already motivated our choice for the scan size, in the following, we investigate how the accuracy of ICP varies with the number of iterations.
Figure~\ref{fig:icp-iter} provides the rotation and translation errors for the experiments depicted in Figure~\ref{fig:icp-a} and Figure~\ref{fig:icp-b}.
More in detail, Figure~\ref{fig:icp-iter-a} shows how the translation error evolves with the number of iterations.
We notice that the translation error tends to converge after about seven iterations and nine iterations for the experiments in Figure~\ref{fig:icp-a} and Figure~\ref{fig:icp-b}, respectively.
Beyond these values, the two curves tend to flatten.
Furthermore, Figure~\ref{fig:icp-iter-b} shows the rotation error.
In this case, we observe a slower convergence trend, but for both experiments, the rotation error decreases below \qty{1}{\degree} for more than nine iterations.
Since we consider it acceptable to have rotation errors below \qty{1}{\degree}, we fix the number of iterations to 10, which is the value used in all our experiments.
While it is true that a dynamic number of iterations is possible, we prefer to use a fixed value to ensure deterministic execution time.

\begin{figure}[t]
    \centering
    \begin{subfigure}[b]{0.99\linewidth}
        \centering
        \includegraphics[width=\linewidth]{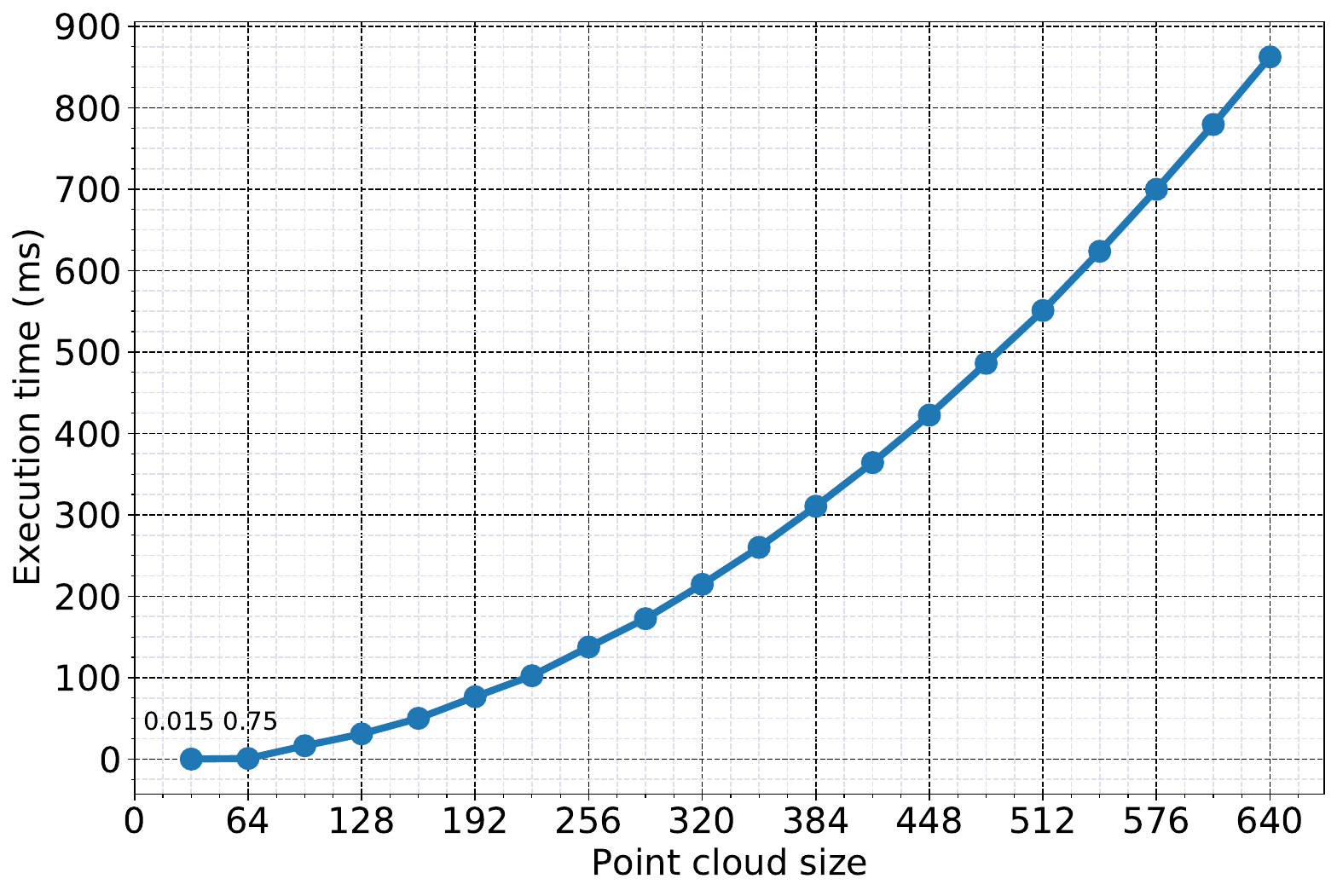}
        \caption{ICP.
        \label{fig:icp-execution-time}}
    \end{subfigure}\hspace{0.04\linewidth}
    \begin{subfigure}[b]{0.99\linewidth}
        \centering
        \includegraphics[width=\linewidth]{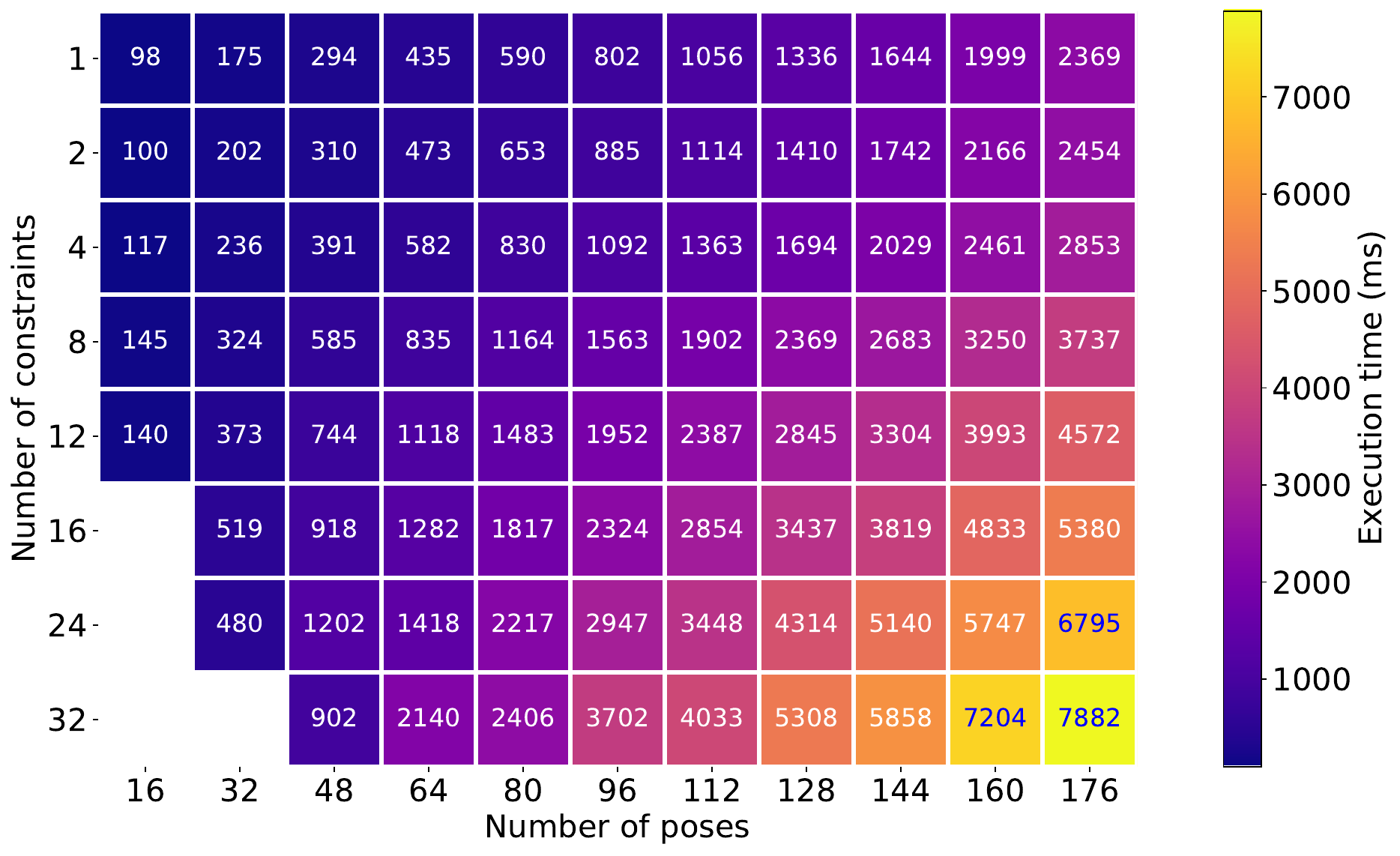}
        \caption{SLAM.
        \label{fig:slam-execution-time}}
    \end{subfigure}
    \caption{Execution times of SLAM and ICP for different configurations.
    \label{fig:execution-time}}
\end{figure}

Next, we evaluate the computation time required to run ICP and SLAM.
Since the execution speed is critical for real-time systems, we analyze how different settings for these algorithms impact the execution time.
For the ICP algorithm, the most critical parameter is the maximum number of 2D points in a scan, which yields to a quadratic increase in the execution time.
Figure~\ref{fig:icp-execution-time} shows how the execution time varies with the size of the scan, using a step of 32 -- the number of 2D points in a scan frame.
As expected, the empirical results also show a purely quadratic dependency.
This is expected because the most dominant computation in ICP is finding the correspondences -- for each point in the first scan, find the closest point in the second scan -- which is implemented as a double for loop.
Matching the empirical measurements presented by Figure~\ref{fig:icp-execution-time} with the quadratic function $0.002\cdot x^{2} + 0.0135 \cdot x - 3.38$ results in a matching over 99\%.
The chosen scan size of 480 points is just below \SI{0.5}{\second} and meets the real-time requirements of the embedded platform.
Since every ICP iteration implies the same computations, obtaining the execution time of one iteration can be performed by dividing the times reported in Figure~\ref{fig:icp-execution-time} by 10.
Moreover, running ICP requires having two scans in the memory, and thus, a scan size of 480 results in a memory consumption of \qty{7680}{\byte}, which is still much smaller than the total amount of \SI{50}{\kilo\byte} available Random Access Memory (RAM).

Next, we provide an evaluation of the execution time of SLAM as a function of both the number of poses and constraints in \cref{fig:slam-execution-time}.
We vary the number of constraints in the range $2^0$ -- $2^5$, which is representative of the number of loop closures typically required in a real-world scenario.
The number of poses is swept in the range of 16 -- 176, with a step of 32.
176 is the largest number of poses that can accommodate 32 constraints without overflowing the available RAM of \SI{50}{\kilo\byte}.
The minimum execution time is \SI{98}{\milli\second} for 16 poses with one constraint, and the maximum is \SI{7882}{\milli\second} for a combination of 176 poses and 32 constraints.
While quite large, even the maximum execution time of almost \SI{8}{\second} is still feasible for a real-world mission.
The only drawback is that the drones have to hover and wait while SLAM is running, which would increase the mission time by \SI{8}{\second} every time SLAM is executed during the mission.

\subsection{Discussion}
We present the limitations of our system as well as possible improvements.
The main goal of our work was to demonstrate that the novel multi-zone ToF sensor can enable new applications for nano-drones, which can generate accurate maps fully onboard due to our optimized implementations of ICP and SLAM.
A notable limitation is the maximum mappable area that our system supports.
As mentioned earlier in this work, this is mainly limited by the maximum amount of poses that the SLAM algorithm can optimize at a time, as well as the inter-pose distance.
With the available onboard RAM of \SI{50}{\kilo\byte}, our system can optimize about 180 poses, which for a \SI{1}{\meter} inter-pose distance corresponds to an area of about  \SI{180}{\meter\squared}.
However, the \SI{1}{\meter} distance for adding a new pose was selected because our testing area is small.
Since the range of the ToF multi-zone sensor is at least \SI{2}{\meter} with less than 5\% invalid pixels \cite{niculescu2022towards}, the inter-pose distance could also be increased to \SI{2}{\meter} which would result in a maximum mappable area four times as large, so about \qty{720}{\meter\squared}.
A more effective way to cope with larger pose graphs is using more onboard RAM memory to optimize large graphs fast enough to meet the real-time requirements.
Emerging low-power embedded MCUs such as multi-core RISC-V SoCs are the perfect candidates for this task as they have a power consumption within hundreds of \qty{}{\milli\watt}, and their parallel capabilities could heavily reduce the execution time of ICP and SLAM, such as in \cite{brunelli2020ultra}. In any case, so far, ICP and SLAM have been executed on workstation-level with general-purpose CPUs, such as the Intel i7-7700HQ featuring a Thermal Design Power (TDP) of \qty{45}{\watt}~\cite{dellenbach2022ct,vizzo2023kissicp}, or onboard with a \qty{7.5}{\watt} computer on module. In \Cref{tab:related}, \cite{huang2022edge} and \cite{zhou2023racer} report a mapping accuracy aligned with this work, but with a power consumption $31\times$ higher, \qty{7.5}{\watt} vs. \qty{240}{\milli\watt} and a flying UAV $52\times$ heavier. Compared with~\cite{dellenbach2022ct,vizzo2023kissicp}, our results achieve a $188\times$ lower power consumption. Regarding the onboard execution time, our optimized ICP is aligned with the SoA of 26-\qty{83}{\milli\second}~\cite{dellenbach2022ct,vizzo2023kissicp} with a point cloud size up to 192, see \Cref{fig:execution-time}, while still respecting the real-time performance requirements for nano-UAV application with an increased number of scans.

Another limitation of our system is the resolution of the ToF sensor. 
While more capable than the other available solutions of its weight and size class, the multi-zone ToF sensor still has a poorer range and resolution than the larger and more power-hungry LiDARs.
As shown in Figure~\ref{fig:scan-frame}, each ToF zone represents a triangle in 2D (or a cone in 3D), whose base increases with the distance magnitude and results in higher uncertainty for the obstacle location.
Therefore, flying the drone closer (i.e., within \qty{2}{\meter}) to the walls or objects is ideal for accuracy-critical processes.
Due to the larger resolution of LiDARs, the zones are narrower, which translates into a higher accuracy that remains in bounds even at large distances.
Moreover, since LiDARs have a higher measurement range than the \qty{4}{\meter} of the multi-zone ToF sensor, they require drones to fly less when mapping larger rooms.

In our work, we perform 2D pose correction and mapping. 
However, since the downward-facing distance sensor directly measures the drone's height, our system is also relevant for 3D mapping applications with uniform ground.
Within this scenario, the drone could change its height during the mission and build an augmented pose structure based on the height provided by the downward-pointing distance sensor and the $x$ and $y$ coordinates provided by SLAM.
In this way, projecting the whole $8 \times 8$ distance matrices from the four ToF sensors in the world frame would result in a 3D point cloud.
This mechanism would require minimal modifications to our system as the scan-matching algorithm would still operate in 2D.
However, for operating in environments with uneven grounds, 3D scan matching is necessary to derive 3D relative constraints between poses.
While the structure of the mapping pipeline would be very similar, more capable MCUs are necessary to enable the heavier real-time computation introduced by an additional degree of freedom.